\documentclass{article} % For LaTeX2e
\usepackage{iclr2027_conference,times}

\usepackage{amsmath,amsfonts,bm}

\def\eqref#1{equation~\ref{#1}}
\def\1{\bm{1}}

\DeclareMathAlphabet{\mathsfit}{\encodingdefault}{\sfdefault}{m}{sl}
\SetMathAlphabet{\mathsfit}{bold}{\encodingdefault}{\sfdefault}{bx}{n}

\usepackage{hyperref}
\usepackage{url}
\usepackage{booktabs}
\usepackage{graphicx}
\usepackage{wrapfig}
\usepackage{needspace}
\usepackage{xcolor}
\usepackage{array}
\usepackage{float}

\title{Language-Forcing: Advancing Humanoid Loco-Manipulation via Semantic Alignment of Whole-Body World Action Models}

\author{%
\begin{tabular}{c}
\textbf{Yanjun Li}\enspace\enspace
\textbf{Jiarui Ye}\enspace\enspace
\textbf{Weiyu Guo}\\
{\mdseries Zixiao He\enspace\enspace Kangjie Ye\enspace\enspace
Dihan Wang\enspace\enspace Yishuo Cheng}\\
{\mdseries Zishang Xiang\enspace\enspace Mingju Gao\enspace\enspace
Hao Tang\thanks{Corresponding author:
\texttt{bjdxtanghao@gmail.com}}}\\
{\mdseries Peking University}\\
{\mdseries Project page:
\url{https://openphyslab.github.io/Language-Forcing/}}
\end{tabular}
}

\iclrfinalcopy

\usepackage{tabularx}
\usepackage{placeins}
\usepackage[most]{tcolorbox}
\definecolor{lfappbgreen}{RGB}{24,103,69}
\definecolor{lfappbblue}{RGB}{0,93,148}
\definecolor{lfappborange}{RGB}{158,84,13}
\newtcolorbox{lfappbpromptbox}[1][]{breakable,colback=black!3,colframe=black!55,boxrule=.5pt,arc=2pt,left=8pt,right=8pt,top=6pt,bottom=6pt,#1}

\newcommand{\lfappbexamplecard}[6]{%
\begin{minipage}[t]{\linewidth}
\small
\setlength{\tabcolsep}{4pt}\renewcommand{\arraystretch}{1.08}
\begin{tabularx}{\linewidth}{@{}>{\bfseries\raggedright\arraybackslash}p{.15\linewidth}X@{}}
\toprule
Image Input & #2 \\[-2pt]
\midrule
Prompt Input & #3 \\
\midrule
Output & {\color{lfappbblue}\textbf{Task progress:} #4}\par\vspace{3pt}
{\color{lfappborange}\textbf{Future prediction:} #5} \\
\bottomrule
\end{tabularx}
\vspace{2pt}\par{\footnotesize\color{black!65}#6}
\end{minipage}}

\usepackage{capt-of}

\begin{document}

% Use a dagger for the corresponding-author footnote.
\begingroup
\renewcommand{\thefootnote}{\ensuremath{\dagger}}
\maketitle
\endgroup
\setcounter{footnote}{0}

\newcommand{\methodname}{\textsc{Language-Forcing}}

% AUTO-GENERATED by scripts/generate_paper_figures.py; do not edit values by hand.
\newcommand{\OursAvgSR}{79.17}
\newcommand{\BaseWAMAvgSR}{68.39}
\newcommand{\StrongestBaselineAvgSR}{69.70}
\newcommand{\StrongestBaselineGain}{9.46}
\newcommand{\LanguageForcingGain}{10.77}
\newcommand{\LFSeventySR}{69.64}
\newcommand{\NoLFHundredSR}{68.39}

\begin{abstract}
World Action Models (WAMs) jointly predict robot actions and future observations. However, their visual prediction objectives do not explicitly distinguish task-relevant transitions from incidental scene changes. For whole-body policies, this distinction matters because task progress depends on coordinated changes in body pose, contact, and object state. We introduce \methodname{}, a training strategy that uses language to emphasize task-relevant changes in the future-visual representations of a whole-body WAM. Given paired current and ground-truth short-horizon future frames, a frozen vision-language model assesses task progress and predicts plausible subsequent state changes. We encode these descriptions with a frozen text encoder and align their embeddings with intermediate, action-conditioned future-visual features, complementing the original visual and action prediction objectives. All auxiliary components are used only during training, introducing no additional inference overhead. On HumanoidArena, \methodname{} achieves an average success rate of $\OursAvgSR\%$, exceeding the strongest competing policy by $\StrongestBaselineGain$ percentage points and the same WAM without language alignment by $\LanguageForcingGain$ percentage points. Across five real-world tasks, language alignment raises average success from $53.33\%$ to $71.33\%$ over the same WAM baseline.
\end{abstract}
\begin{figure*}[ht!]
    \centering
    \includegraphics[width=0.96\linewidth]{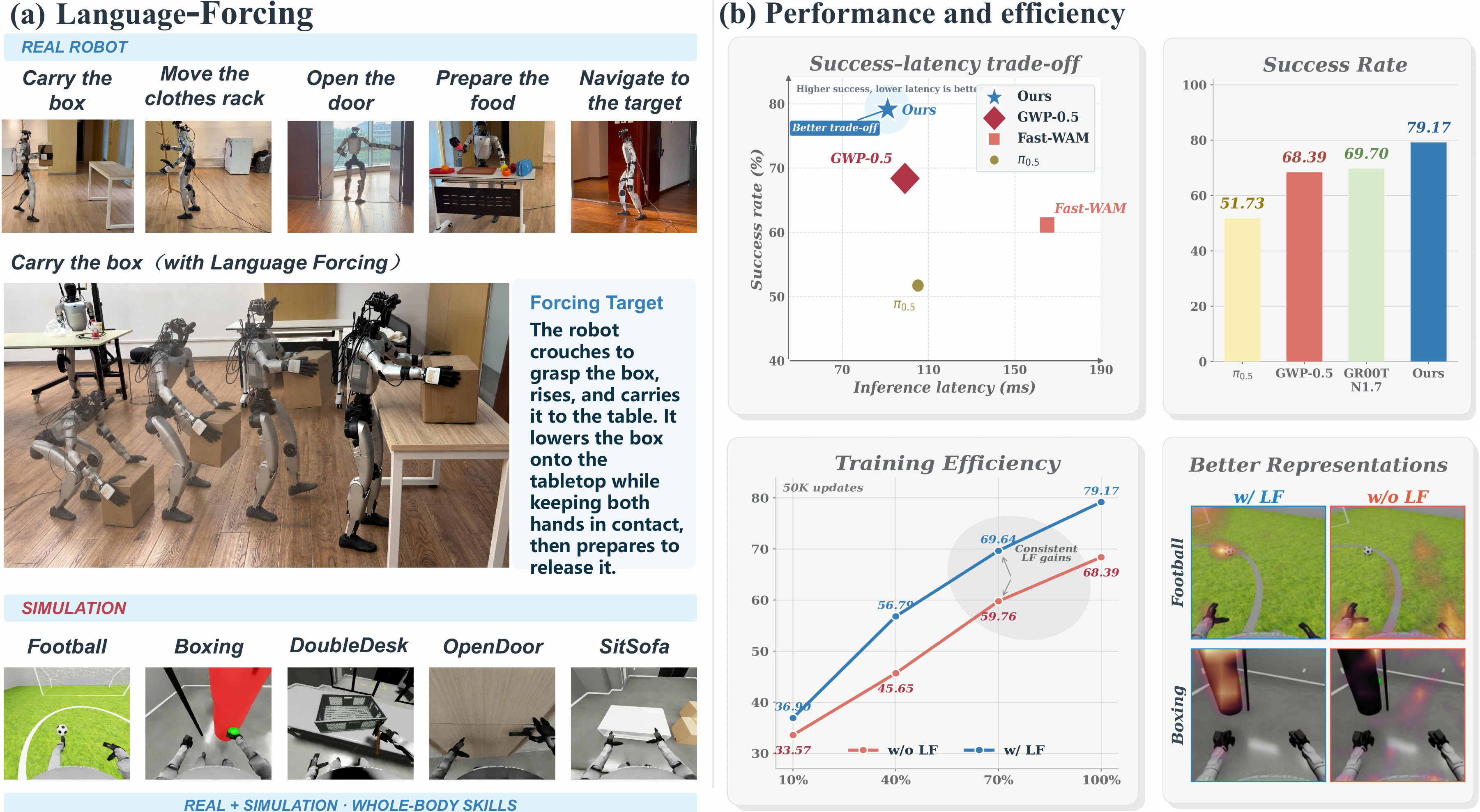}
    \vspace{-0.13in}
    \caption{
    \textbf{\methodname{} improves whole-body WAMs without additional inference overhead.}
    \textbf{(a)} Whole-body tasks in simulation and on a real humanoid, with an example language target for box transport. These targets describe task progress and likely subsequent state changes and are used only during training.
    \textbf{(b)} Policy success rates, inference latency, and data efficiency. The spatial attribution maps highlight the football and punching bag with Language Forcing, illustrating the task-relevant focus examined further in Figure~\ref{fig:spatial_focus}.
    }
    \label{fig:teaser}
\end{figure*}

\section{Introduction}

World models learn predictive state representations for control~\citep{finn2017visualforesight,hafner2019planet}, and recent video-based policies and WAM-style formulations increasingly couple action generation with future state prediction~\citep{du2023learninguniversalpoliciestextguided,hu2025videopredictionpolicygeneralist,ye2026worldactionmodelszeroshot,ma2026dit4ditjointlymodelingvideo}. In parallel, vision-language-action models improve instruction following and generalization, with growing coverage of whole-body humanoid interaction~\citep{brohan2023rt2visionlanguageactionmodelstransfer,octomodelteam2024octoopensourcegeneralistrobot,jiang2025wholebodyvlaunifiedlatentvla,nvidia2025gr00tn1openfoundation}. Despite this progress, whole-body action selection depends on coupled robot--object states: contact enables manipulation and stance adjustments enable reaching~\citep{sentis2006wholebody}. Predictive representations must therefore emphasize the state changes that actually advance the task.

Visual prediction offers dense supervision but can overemphasize large, task-irrelevant changes (e.g., viewpoint shifts) while underweighting subtle but action-critical events (e.g., first contact with a handle). The resulting representation may therefore be misaligned with what the policy needs to decide the next action. Since the task instruction specifies the goal rather than per-transition progress, additional guidance is needed to focus predictive features on task advancement; closed-loop results also suggest visual fidelity alone is not sufficient~\citep{zhang2026worldinworld}.

Language can make this distinction explicit: grasping a handle
identifies task progress, while predicting that the door will open
describes a possible subsequent outcome.
The task instruction remains the same across these stages, whereas
the desired language target changes with the observed transition.
Prior work uses language to learn robot representations and rewards
\citep{karamcheti2023languagedrivenrepresentationlearningrobotics,
ma2023livlanguageimagerepresentationsrewards}
or introduces spatial knowledge into policies
\citep{qu2025spatialvla,zhang2026spatialtoactions}.
Spatial Forcing aligns visual features with pretrained geometric
representations~\citep{li2025spatialforcingimplicitspatial},
while WorldScape Policy 2.0 supervises WAM planning representations
with event descriptions~\citep{su2026worldscapepolicy20empowering}.
% Add the corresponding citation for WorldScape Policy 2.0.
We focus on the action-conditioned future-visual representation.
These features are learned jointly with action generation and
conditioned on the action trajectory, making them a natural point
for relating predicted physical changes to their task-level meaning.
Our insight is that assessments of task progress and predictions
of subsequent states can guide these features toward changes
relevant to action generation.
Visual and language supervision are complementary: one captures
scene dynamics, while the other relates them to task progress
and likely next steps.

We introduce \methodname{}, a training strategy for whole-body WAMs.
A frozen vision-language model receives the task context and paired
current and future observations, assesses task progress, and describes
a plausible continuation beyond the observed endpoint.
A frozen UMT5 encoder~\citep{chung2023unimaxfairermoreeffective}
encodes its response for alignment with intermediate future-visual
features conditioned on the scene, robot state, and action trajectory.
This supervision complements the original visual and action objectives.
All auxiliary components are used only during training, leaving the
deployed action-generation path unchanged without additional
inference computation.

On HumanoidArena
\citep{wang2026humanoidarenabenchmarkingegocentrichierarchical},
\methodname{} achieves $79.17\%$ average success across seven tasks,
exceeding the strongest competing policy by $9.46$ percentage points.
Matched WAM comparisons and five real-world tasks further
demonstrate its effectiveness.

Our contributions are threefold:
\begin{itemize}
    \item We introduce Language Forcing, which uses a VLM's assessments
    of task progress and predictions of future states to supervise
    the intermediate visual features used by a WAM to predict
    future observations.

    \item We incorporate Language Forcing into a whole-body humanoid
    WAM for modeling whole-body actions, without additional computational overhead at test time.

    \item We achieve state-of-the-art performance on HumanoidArena
    and demonstrate the effectiveness of Language Forcing through
    a matched comparison with the same WAM trained without
    language supervision.
\end{itemize}

\section{Related Work}

\paragraph{Vision-language-action and world-action models.}
Generalist robot policies learn instruction-conditioned actions across tasks and embodiments, drawing on large-scale demonstrations and pretrained vision-language representations~\citep{brohan2023rt1roboticstransformerrealworld,driess2023palmeembodiedmultimodallanguage,brohan2023rt2visionlanguageactionmodelstransfer,octomodelteam2024octoopensourcegeneralistrobot,kim2024openvlaopensourcevisionlanguageactionmodel}.
Diffusion and flow-based policies support continuous action generation~\citep{chi2023diffusionpolicyvisuomotor,black2026pi0visionlanguageactionflowmodel,intelligence2025pi05visionlanguageactionmodelopenworld}, while related work studies data diversity, few-shot adaptation, cross-embodiment learning, and instruction tuning~\citep{lin2025datascalinglawsimitation,vosylius2025instantpolicy,zheng2026xvlasoftpromptedtransformer,yang2026instructvlavisionlanguageaction}.
Predictive visuomotor methods use future videos or visual features for planning and action learning~\citep{du2023learninguniversalpoliciestextguided,tian2025predictiveinversedynamics,hu2025videopredictionpolicygeneralist,zhou2025dinowmworldmodelspretrained}.
World-action models further couple future prediction with action generation~\citep{kim2026cosmospolicyfinetuningvideo,bi2025motusunifiedlatentaction,ma2026dit4ditjointlymodelingvideo,ye2026worldactionmodelszeroshot}, alongside efforts to evaluate world models through closed-loop control~\citep{zhang2026worldinworld}.
Recent approaches investigate action-centered architectures and efficient predictive representations~\citep{ye2026gigaworldpolicyefficientactioncenteredworldaction,yuan2026fastwamworldactionmodels,luo2026beingh07latentworldactionmodel,chen2026lawamlatentworldaction}, while MotionWAM and $\omega$-0 extend predictive learning to humanoid whole-body actions~\citep{zheng2026motionwamfoundationworldaction,li2026omega0latentpredictiveworld}.
\paragraph{Humanoid whole-body policies.}
Humanoid policies have progressed from whole-body motion tracking~\citep{he2024omnih2ouniversaldexteroushumantohumanoid,he2025hoverversatileneuralwholebody,luo2026sonicsupersizingmotiontracking} toward promptable and language-guided control for coordinated movement and task execution~\citep{li2026bfmzerohumanoidcontrol,li2026languagetolocomotion}.
GR00T N1 and WholeBodyVLA extend this direction to generalist humanoid policies and loco-manipulation~\citep{nvidia2025gr00tn1openfoundation,jiang2025wholebodyvlaunifiedlatentvla}.
RobotArena $\infty$ and HumanoidArena provide large-scale policy evaluation and egocentric tasks requiring coordination of the legs, torso, arms, and hands, respectively~\citep{jangir2026robotarenainfinity,wang2026humanoidarenabenchmarkingegocentrichierarchical}.
\paragraph{Language and representation supervision for policy learning.}
Visual pretraining and language supervision provide transferable representations for robot learning~\citep{xiao2022maskedvisualpretraining,nair2022r3muniversalvisualrepresentation,karamcheti2023languagedrivenrepresentationlearningrobotics,ma2023livlanguageimagerepresentationsrewards}.
Spatially informed policies further incorporate explicit spatial encodings, 3D feature alignment, or pretrained spatial features~\citep{qu2025spatialvla,li2025spatialforcingimplicitspatial,zhang2026spatialtoactions}.
Being-H0.7 transfers future-informed representations into a deployable policy~\citep{luo2026beingh07latentworldactionmodel}, while WorldScape Policy 2.0 supervises planning representations with event captions~\citep{su2026worldscapepolicy20empowering}.
Language-Forcing aligns action-conditioned future-visual features with language representations describing observed short-horizon state transitions, using this auxiliary supervision only during training.
\section{Method}
\label{sec:method}

\subsection{Preliminaries}
\label{sec:preliminaries}

\paragraph{World Action Models.}
World Action Models (WAMs) jointly model robot actions and future
visual observations.
Our base policy follows an action-centered WAM with separate visual
and action experts
\citep{ye2026gigaworldpolicyefficientactioncenteredworldaction}.
Given the current egocentric image $I_t$, robot state $s_t$,
and task instruction $c_t$, the policy outputs whole-body actions.
During training, it also predicts future video latents over
the same action horizon.

The model organizes its inputs into four token groups.
State tokens $S$ encode the robot state, and reference-visual
tokens $R$ encode the current image.
Action tokens $U$ and future-visual tokens $F$ represent
the noisy action sequence and future video latents, respectively.
A mixture-of-transformers backbone processes these tokens
with a directional attention mask.
Future-visual queries attend to $S$, $R$, $U$, and $F$,
so future visual prediction is conditioned on the actions.
Action queries attend only to $S$, $R$, and $U$.
Both experts receive task-instruction embeddings through cross-attention.
The action branch can therefore run without future-visual tokens
at inference.

\paragraph{VLM teacher.}
We use a frozen Gemma 4 12B
\citep{gemmateam2026gemma4technicalreport}
as the vision-language teacher.
It receives paired current and future observations
together with the task context.
And text outputs are used to construct language targets
for training the WAM.
This process is performed offline, and the teacher is omitted
at inference.

\subsection{Motivation}
\label{sec:motivation}

\paragraph{Challenge.}
Future visual prediction requires modeling changes in robot pose,
object position, contact, camera motion, and appearance.
However, these changes differ in their importance for the task.
A shift in viewpoint may affect much of the image, whereas
establishing contact with an object may produce only a small
visual change.
The visual flow loss does not explicitly distinguish between them
according to their relevance to task progress.
The task instruction specifies the goal but does not describe
what has been completed in a particular training sample.
WAMs therefore receive supervision for the future observation
without an explicit assessment of task progress.

\paragraph{Observation.}
Current and future frames show how the robot and surrounding objects change during a training sample.
Together with the task instruction, visual input allow a VLM to assess
task progress and predict future states relevant to the task.
For example, a powerful VLM can identify that the robot has grasped
a handle and predict that the door will open.
The resulting text describes the stage reached by the task
and what is expected to happen next.

We use this language supervision to train the intermediate visual
features that the WAM uses to predict future observations.
These features are computed from the current image, robot state,
and action tokens.
Aligning them with the language targets encourages the model
to learn which changes in the predicted future are important
for completing the task.
Visual loss is retained to train future video prediction.

\begin{figure*}[t]
    \centering
    \includegraphics[width=\textwidth]{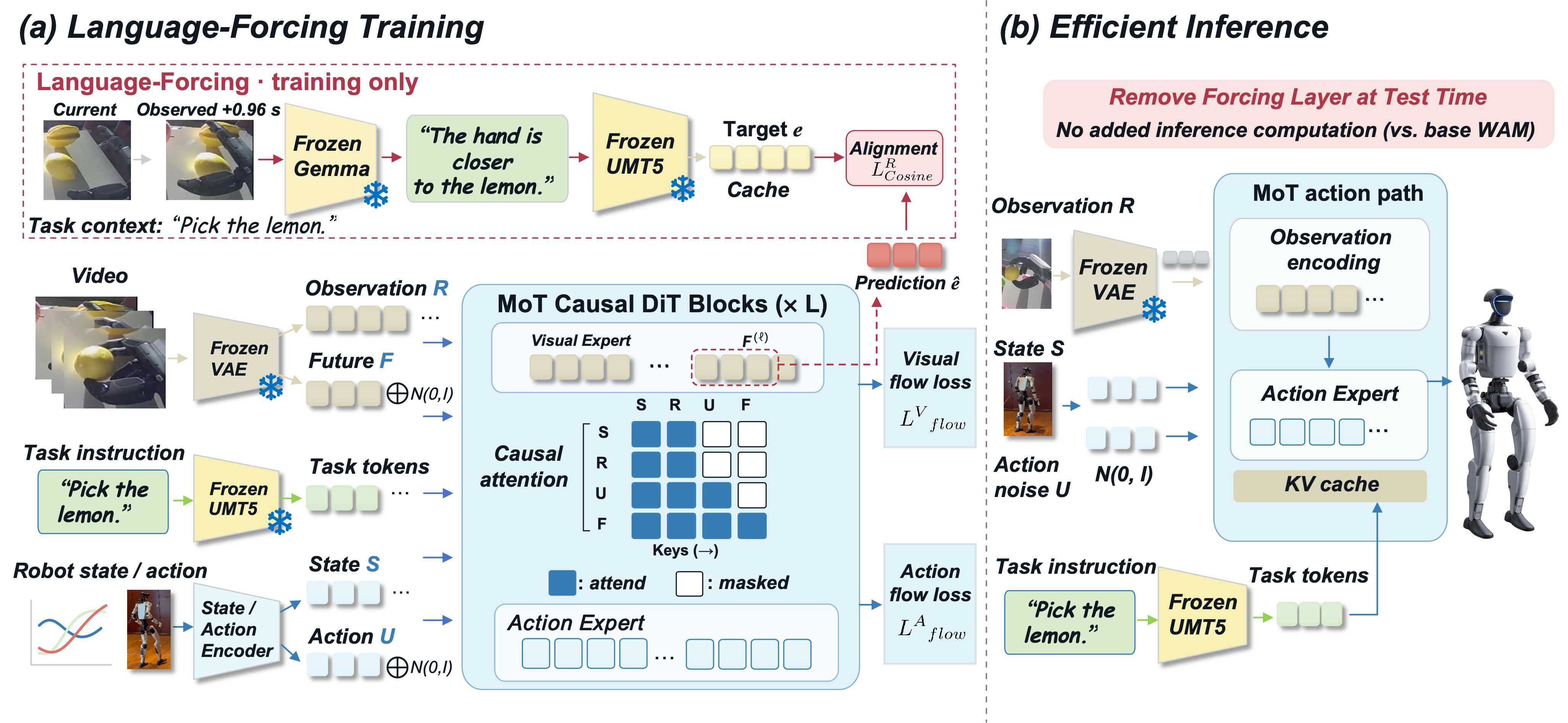}
    \caption{\textbf{Overview of \methodname{}.}
    (a) Training aligns intermediate future-visual features with VLM-generated language targets.
    (b) The auxiliary branch is removed at inference, preserving the action expert.}
    \label{fig:main}
\end{figure*}

Figure~\ref{fig:main} illustrates the training and inference
procedures of \methodname{}.

\subsection{Language Forcing}
\label{sec:language_forcing}

\paragraph{Language supervision.}
For each training sample, we pair the current observation $I_t$
with future observation $I_{t+48}$.
The two frames are separated by $\Delta=0.96$ seconds
in the $50$ Hz trajectory.
We place them side by side, with the current frame on the left.
The frozen VLM receives the paired image, the elapsed time
$\Delta$, and the task instruction $c_t$.
Its output $y_t$ assesses task progress and predicts future states
relevant to the task.

A frozen UMT5 encoder $E_{\mathrm{txt}}$
\citep{chung2023unimaxfairermoreeffective}
encodes $y_t$ into token features.
We average the non-padding token features to obtain the language target:
\begin{equation}
    e_t
    = \frac{1}{|\mathcal J_t|}
      \sum_{j\in\mathcal J_t}
      \bigl[E_{\mathrm{txt}}(y_t)\bigr]_j,
    \label{eq:text_pooling}
\end{equation}
where $\mathcal J_t$ contains the indices of non-padding tokens.
The resulting embedding $e_t\in\mathbb{R}^{4096}$ is used
for feature alignment.
It is separate from the task-instruction embeddings, which condition
the WAM during both training and inference.
We generate the text and compute its embedding offline,
then cache both for training.

\paragraph{Feature alignment.}
We apply language supervision to the intermediate visual tokens
used to predict future observations.
These tokens attend to the current image, robot state, and action tokens.
Let $F_t^{(\ell)}\in\mathbb{R}^{N_f\times d_v}$ denote their
hidden features after transformer block $\ell$,
where $N_f$ is the number of future-visual tokens and $d_v$
is the feature dimension.
Tokens from the current image are excluded.
We average the future-visual features and project them
to the same dimension as the language target:
\begin{equation}
    \widehat e_t
    = P_\phi\!\Bigl(
        \frac{1}{N_f}
        \sum_{i=1}^{N_f} F_{t,i}^{(\ell)}
      \Bigr).
    \label{eq:visual_pooling}
\end{equation}
The trainable projector $P_\phi$ consists of layer normalization,
a linear layer, GELU, and a second linear layer.
We align the projected features with the language target
using a cosine loss:
\begin{equation}
    \mathcal L_{\mathrm{LF}}
    = 1 - \frac{1}{B}
      \sum_{b=1}^{B}
      \cos\!\bigl(\widehat e_b,e_b\bigr),
    \label{eq:language_forcing}
\end{equation}
where $B$ is the batch size, $b$ indexes training samples,
and $\cos(\cdot,\cdot)$ denotes cosine similarity.
Language loss supervises the visual features, while the visual flow loss supervises future video prediction.

\paragraph{Training objective.}
We add the language loss to the original WAM objective:
\begin{equation}
    \mathcal L
    = \mathcal L_{\mathrm{WAM}}
    + \lambda_{\mathrm{LF}}\,\mathcal L_{\mathrm{LF}},
    \label{eq:complete_objective}
\end{equation}
where $\lambda_{\mathrm{LF}}$ controls the weight of language supervision.
The base objective $\mathcal L_{\mathrm{WAM}}$ contains the action flow
and visual flow losses.

At each training step, we add noise to the action targets and future
video latents at independently sampled noise levels.
A single forward pass through the WAM produces the action and visual
flow predictions.
We also extract $F_t^{(\ell)}$ from block $\ell$ to compute
the language loss in Equation~\ref{eq:language_forcing}.
The combined loss updates the WAM and the projector $P_\phi$.
The trainable WAM components include the visual and action experts,
output heads, and humanoid input and output adapters.
The visual autoencoder, VLM teacher, and target UMT5 encoder remain frozen.

\paragraph{Inference.}
At inference, policy receives the current image
$I_t$, robot state $s_t$, and task instruction.
And outputs whole-body actions.
The language-alignment components are omitted at deployment,
so Language Forcing does not introduce additional computational
overhead at inference.

\begingroup % Scope experiment float spacing to the main paper.

\setlength{\textfloatsep}{10pt plus 2pt minus 2pt}
\setlength{\floatsep}{8pt plus 2pt minus 2pt}
\setlength{\abovecaptionskip}{6pt}
\section{Experiments}
\label{sec:experiments}

We evaluate \methodname{} on HumanoidArena and five real-world tasks.
The ablations examine the content of the language targets, their
correspondence with the observations, and the future visual information
used during training.

\subsection{Experimental Setup}
\label{sec:experimental_setup}

\paragraph{Benchmark and metric.}
HumanoidArena \citep{wang2026humanoidarenabenchmarkingegocentrichierarchical}
evaluates humnanoid whole-body policies on seven tasks.
The human-object interaction(HOI) tasks including Football, DoubleDesk, and P\&PBox;
the human-scene interaction(HSI) tasks including OpenDoor, SitSofa, Boxing, and VisNavi.
All policies use SONIC for motion tracking during evaluation.
We evaluate each task under the base mode and three out-of-distribution
(OOD) modes: lighting changes (visual), altered appearance or distractors (semantic), and object positions outside the training range (execution).
Each mode uses three rollout seeds and 20 trials per seed,
giving 240 trials per task and 1,680 trials in total.
We report success rate (SR), with equal weight for the four settings,
and the unweighted mean across tasks (AVG).
We report the success rate (SR) over all 240 trials for each task and average success rate across all tasks.
% Define the four evaluation settings in the implementation appendix.

\paragraph{Baselines.}
We compare with established visuomotor policies, including
ACT \citep{zhao2023learningfinegrainedbimanual},
Diffusion Policy (DP) \citep{chi2023diffusionpolicyvisuomotor},
and a flow-matching policy (FM) \citep{lipman2023flowmatchinggenerativemodeling}.
We also evaluate recent vision-language-action (VLA) models:
$\pi_{0.5}$ \citep{intelligence2025pi05visionlanguageactionmodelopenworld}
and GR00T N1.7 \citep{nvidia2025gr00tn1openfoundation}.
Representative WAM baselines include
LingBot-VA \citep{li2026causalworldmodelingrobot},
Fast-WAM \citep{yuan2026fastwamworldactionmodels},
and GigaWorld-Policy 0.5
\citep{ye2026gigaworldpolicyefficientactioncenteredworldaction}.
GigaWorld-Policy 0.5 also serves as the base WAM for \methodname{}.
ACT, DP, FM, and $\pi_{0.5}$ are reevaluated under the same protocol.

\paragraph{Training.}
Training demonstrations contains 200 successful episodes per task, for 1,400 episodes in total.
All methods use same training data and action interface.
We match the observation sources, initial configurations, rollout seeds,
and success criteria within each evaluation setting.
Each model is trained for 50K updates using its architecture specific
preprocessing and optimization settings.

We initialize \methodname{} from GigaWorld-Policy 0.5 and adapt its action
interface to the whole-body representation in Section~\ref{sec:preliminaries}.
The WAM receives $320\times384$ images and five frames at offsets
$\{0,12,24,36,48\}$ in the 50 Hz trajectory.
The models with and without Language Forcing use the same initialization,
humanoid adapters, training data, and base optimization settings;
the added language loss is the only difference.
We use the final 50K step exponential moving average (EMA) weights for evaluation.
All ablations retain this training budget.
Data partitions and model-specific settings are provided in
Appendix~\ref{sec:implementation_details}.

\subsection{Policy Performance}
\label{sec:main_results}

% AUTO-GENERATED by scripts/generate_paper_figures.py; do not edit values by hand.
\begin{table}[t]
    \centering
    \caption{Success rate (SR, \%) on HumanoidArena. AVG is the unweighted
    mean across seven tasks. Bold and underlined entries indicate the best
    and second-best distinct scores in each column, respectively.We denote Double Desk and Open Door by DouDesk and OpDoor, respectively.}
    \label{tab:main_results}
    \small
    \setlength{\tabcolsep}{2pt}
    \renewcommand{\arraystretch}{1.05}
    \begin{tabular*}{\linewidth}{@{\extracolsep{\fill}}lrrrrrrrr@{}}
        \toprule
        & \multicolumn{3}{c}{Human--Object Interaction}
        & \multicolumn{4}{c}{Human--Scene Interaction} & \\
        \cmidrule(lr){2-4}\cmidrule(lr){5-8}
        Method & Football & \shortstack{DouDesk} & P\&PBox & \shortstack{OpDoor} & SitSofa & Boxing & VisNavi & AVG \\
        \midrule
        ACT~\citep{zhao2023learningfinegrainedbimanual} & 17.92 & 19.17 & 57.92 & 79.17 & 74.58 & 58.33 & 35.00 & 48.87 \\
        DP~\citep{chi2023diffusionpolicyvisuomotor} & 46.67 & 37.92 & \underline{76.25} & 86.25 & \underline{79.17} & \textbf{77.50} & 25.42 & 61.31 \\
        FM~\citep{lipman2023flowmatchinggenerativemodeling} & 14.58 & 39.58 & 74.58 & 71.25 & 16.25 & 70.83 & 40.00 & 46.73 \\
        $\pi_{0.5}$~\citep{intelligence2025pi05visionlanguageactionmodelopenworld} & 11.25 & 42.08 & 72.92 & 67.92 & 74.58 & 68.75 & 24.58 & 51.73 \\
        Fast-WAM~\citep{yuan2026fastwamworldactionmodels} & 46.25 & 45.00 & 73.75 & 85.42 & 74.58 & 67.92 & 35.00 & 61.13 \\
        LingBot-VA~\citep{li2026causalworldmodelingrobot} & 51.67 & 41.25 & \textbf{77.92} & 84.58 & 77.08 & 72.92 & 52.08 & 65.36 \\
        GR00T N1.7~\citep{nvidia2025gr00tn1openfoundation} & \underline{63.75} & 49.58 & 73.33 & 82.08 & 76.67 & \underline{76.67} & \underline{65.83} & \underline{69.70} \\
        GigaWorld-Policy 0.5~\citep{ye2026gigaworldpolicyefficientactioncenteredworldaction} & 62.08 & \underline{52.50} & 65.83 & \underline{90.83} & 74.17 & 73.33 & 60.00 & 68.39 \\
        \midrule
        \methodname{} (Ours) & \textbf{83.33} & \textbf{62.92} & 74.58 & \textbf{97.08} & \textbf{87.50} & 74.58 & \textbf{74.17} & \textbf{79.17} \\
        \bottomrule
    \end{tabular*}
\end{table}

\paragraph{Overall performance.}
\methodname{} achieves the highest average SR of $79.17\%$
(Table~\ref{tab:main_results}), exceeding the strongest baseline,
GR00T N1.7, by $9.46$ percentage points.
The results cover both object interaction and scene interaction,
which require different combinations of locomotion, body positioning,
and manipulation.

\paragraph{Effect of Language Forcing.}
Compared with the same WAM without Language Forcing, our method raises average SR by $10.77$ percentage points.
Language-Forcing improves performance on tasks involving object motion, such as Football and DoubleDesk, as well as body positioning in SitSofa and navigation in VisNavi.
The base WAM already learns to predict future observations.
Adding task-relevant semantic supervision and future-state predictions as
supervision for its visual features further improves the policy,
without changing its inference procedure.
The ablations below examine what makes this language supervision useful.

\begin{figure}[tbp]
    \centering
    \includegraphics[width=0.98\textwidth]{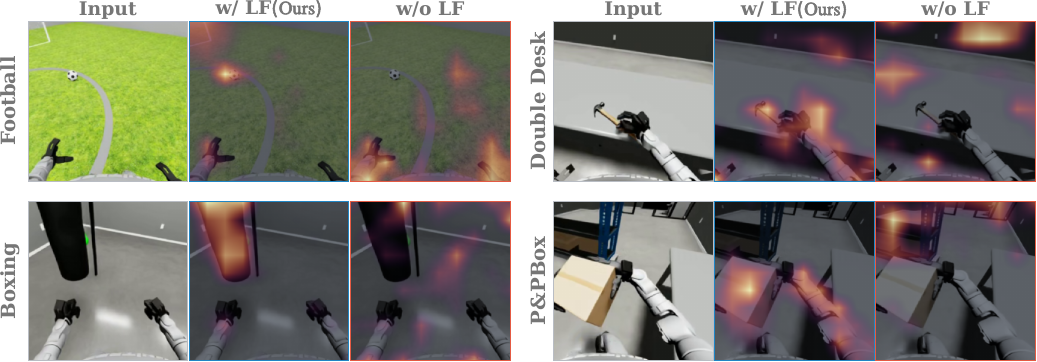}
    \caption{\textbf{Task-grounded spatial attribution with Language-Forcing.}
    Input images and WAM attribution maps with and without Language Forcing (LF).
    LF emphasizes the football, punching bag, and hand--object interactions
    in DoubleDesk and P\&PBox, while baseline responses are more dispersed.
    Football and Boxing also appear in Figure~\ref{fig:teaser}(b).}
    \label{fig:spatial_focus}
\end{figure}

\paragraph{Task-relevant spatial focus.}
Figure~\ref{fig:spatial_focus} expands the representation visualization in
Figure~\ref{fig:teaser}(b). With Language Forcing, responses focus on the ball
in Football and the punching bag in Boxing, and on the manipulated object and
interacting hand in DoubleDesk and P\&PBox. Without Language Forcing,
responses are more dispersed over background regions. These qualitative
examples support the task-relevant focus of the learned visual features,
complementing the performance gains in Table~\ref{tab:main_results}.

\subsection{Language-Forcing Design Analysis}
\label{sec:ablations}

We first investigate different language supervision targets,
their correspondence with observed transitions, and the frames
used for visual prediction. Then we study the effects of alignment
depth and language-loss weight, and compare performance with
different numbers of demonstrations.

\begin{wrapfigure}{r}{0.48\linewidth}
    \centering
    \includegraphics[width=\linewidth]{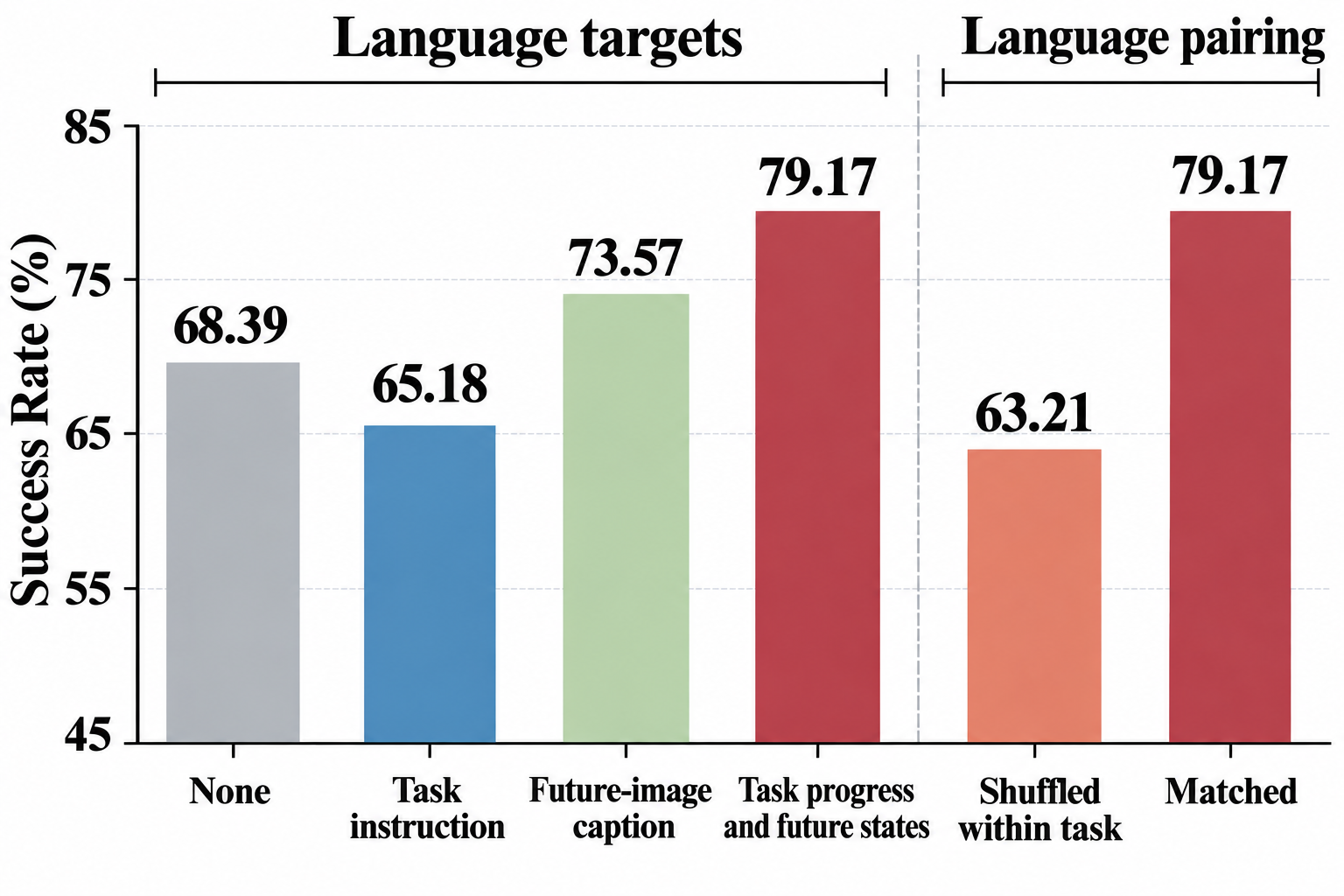}
    \setlength{\abovecaptionskip}{3pt}
    \setlength{\belowcaptionskip}{0pt}
    \caption{Language supervision on HumanoidArena.
    Left: alignment targets with identical VLM image inputs.
    Right: matched versus within-task shuffled descriptions.
    WAM training images are fixed.}
    \label{fig:language_supervision_ablation}
\end{wrapfigure}
\noindent
\textbf{Task relevant semantics versus image captions.}\enspace
We compare task instructions, captions of the future image,
and the descriptions used by Language Forcing as alignment targets
(Figure~\ref{fig:language_supervision_ablation}, left).
For both captioning and Language Forcing, the VLM receives
the current and future images.
The caption prompt asks for a description of the future scene.
Language Forcing asks the VLM to assess progress and predict future states.
The WAM's visual targets and task-instruction conditioning remain unchanged.
Captions improve SR over training without language alignment,
while Language Forcing adds a further $5.60$ percentage points.
Task instructions give the lowest SR among the three language targets.
With the same observations available to the VLM, language about
task progress and future states provides more effective supervision
than a description of the future scene alone.
\par

\paragraph{Matching language to observations.}
We shuffle language descriptions across trajectories of the same task
(Figure~\ref{fig:language_supervision_ablation}, right).
The descriptions remain relevant to the task, but are no longer paired
with the observations used to produce them.
Matching each description to its original observations improves SR
by $15.95$ percentage points over shuffling.
Thus, descriptions of the same task cannot be used interchangeably
for language alignment.
The language needs to reflect the robot actions and object states of the corresponding
observations, where the robot and objects may be at different
stages of the task.

\paragraph{Predicting future frames.}
We keep the language descriptions fixed and change the images used
for future video prediction.
Every setting uses five image inputs, with frames repeated when necessary.
Repeating the current image in all five positions yields $65.18\%$ SR.
Including the current image, one intermediate frame, and the endpoint
increases SR to $72.98\%$, while using three intermediate frames raises
it to $79.17\%$, with the current image and endpoint unchanged.
The latter comparison isolates the benefit of including more images
of the motion before the endpoint.
Language provides task-relevant semantic supervision and describes future states,
while the intermediate frames show how the robot and objects move.
The improvement from these frames explains why future video prediction
remains useful alongside language alignment.
The frame offsets and repeated inputs are listed in
Appendix~\ref{sec:implementation_details}.

With the language descriptions and training images fixed,
we next examine where to apply alignment and how to weight the language loss.

\begin{table}[t]
    \begin{minipage}[t]{0.57\linewidth}
            \centering
    \small
    \setlength{\abovecaptionskip}{2pt}
    \setlength{\belowcaptionskip}{4pt}

    \captionof{table}{Independent sweeps of alignment depth (left) and loss weight (right). SR (\%).}
    \label{tab:ablation-alignment-depth}
    \label{tab:ablation-loss-weight}

    \setlength{\tabcolsep}{2.5pt}
    \renewcommand{\arraystretch}{1.05}

    \begin{tabular*}{\linewidth}{@{\extracolsep{\fill}}cr@{\hspace{16pt}}cr@{}}
        \toprule
        Block $\ell$ & SR & $\lambda_{\mathrm{LF}}$ & SR \\
        \cmidrule(r){1-2}\cmidrule(l){3-4}
        1  & 73.99 & 0   & 68.39 \\
        8  & 74.23 & 0.5 & 74.40 \\
        15 & 75.48 & 1.0 & \textbf{79.17} \\
        22 & \textbf{79.17} & 1.5 & 73.39 \\
        29 & 74.46 & 2.0 & 68.75 \\
        \bottomrule
    \end{tabular*}

    \end{minipage}\hfill
    \begin{minipage}[t]{0.39\linewidth}
            \centering
    \small
    \setlength{\abovecaptionskip}{2pt}
    \setlength{\belowcaptionskip}{4pt}

    \captionof{table}{Training data. SR (\%) with 50K updates.
    Underlining marks the $70\%$ LF and $100\%$ baseline comparison.}
    \label{tab:ablation-data-efficiency}

    \setlength{\tabcolsep}{4pt}
    \renewcommand{\arraystretch}{1.05}

    \begin{tabular*}{\linewidth}{@{\extracolsep{\fill}}crr@{}}
        \toprule
        Data & w/o LF & w/ LF \\
        \midrule
        $10\%$ & 33.57 & 36.90 \\
        $40\%$ & 45.65 & 56.79 \\
        $70\%$ & 59.76 & \underline{69.64} \\
        $100\%$ & \underline{68.39} & \textbf{79.17} \\
        \bottomrule
    \end{tabular*}

    \end{minipage}
\end{table}

\noindent
\textbf{Alignment layer.}\enspace
We apply language alignment at five blocks of the visual expert,
keeping $\lambda_{\mathrm{LF}}=1.0$
(Table~\ref{tab:ablation-alignment-depth}, left).
All five improve upon the WAM without Language Forcing,
and block 22 achieves the highest SR.
Moving alignment from block 22 to block 29 reduces performance,
so a layer closer to the visual output is not necessarily a better
choice for language supervision.
We use block 22 in \methodname{}; block indices are zero-based.

\paragraph{Language-loss weight.}
With alignment at block 22, we vary the weight of the language loss
relative to the visual and action losses
(Table~\ref{tab:ablation-loss-weight}, right).
SR peaks at $\lambda_{\mathrm{LF}}=1.0$ and decreases at larger weights.
At $\lambda_{\mathrm{LF}}=2.0$, performance approaches the WAM baseline.
Language alignment therefore benefits action modeling at a moderate weight,
but increasing its contribution to the training objective does not
produce a larger gain.
We use $\lambda_{\mathrm{LF}}=1.0$ in \methodname{}.
\par

\noindent
\textbf{Data efficiency.}\enspace
We compare the WAM with and without Language Forcing using the same
subsets of demonstrations and 50K training updates
(Table~\ref{tab:ablation-data-efficiency}).
Language Forcing improves SR at every training-set size.
With $70\%$ of the demonstrations, the WAM with Language Forcing
reaches $69.64\%$ SR, compared with $68.39\%$ for the WAM trained
on all demonstrations without language alignment.
This corresponds to comparable performance with $30\%$ fewer
demonstrations in this experiment.
The gains also persist as more demonstrations become available.
From $40\%$ to $100\%$ of the training set, Language Forcing maintains
an improvement of around 10 percentage points.
The benefit therefore extends beyond training with very few demonstrations:
adding more robot data does not remove the gain from language supervision.
The VLM extracts task-relevant semantic information from the recorded images,
providing additional supervision without collecting new robot demonstrations.
\par

\subsection{Real-World Experiments}
\label{sec:real_world_experiments}

We evaluate \methodname{} on five real-world tasks and compare with
the same WAM without Language Forcing, GR00T N1.7, and Fast-WAM.
The comparison with the same WAM examines whether language supervision
improves physical task completion under the same training and
deployment settings.
The robot hardware and deployment workstation are described in
Appendix~\ref{sec:implementation_details}.

\begin{figure}[tbp]
    \centering
    \includegraphics[width=0.86\linewidth]{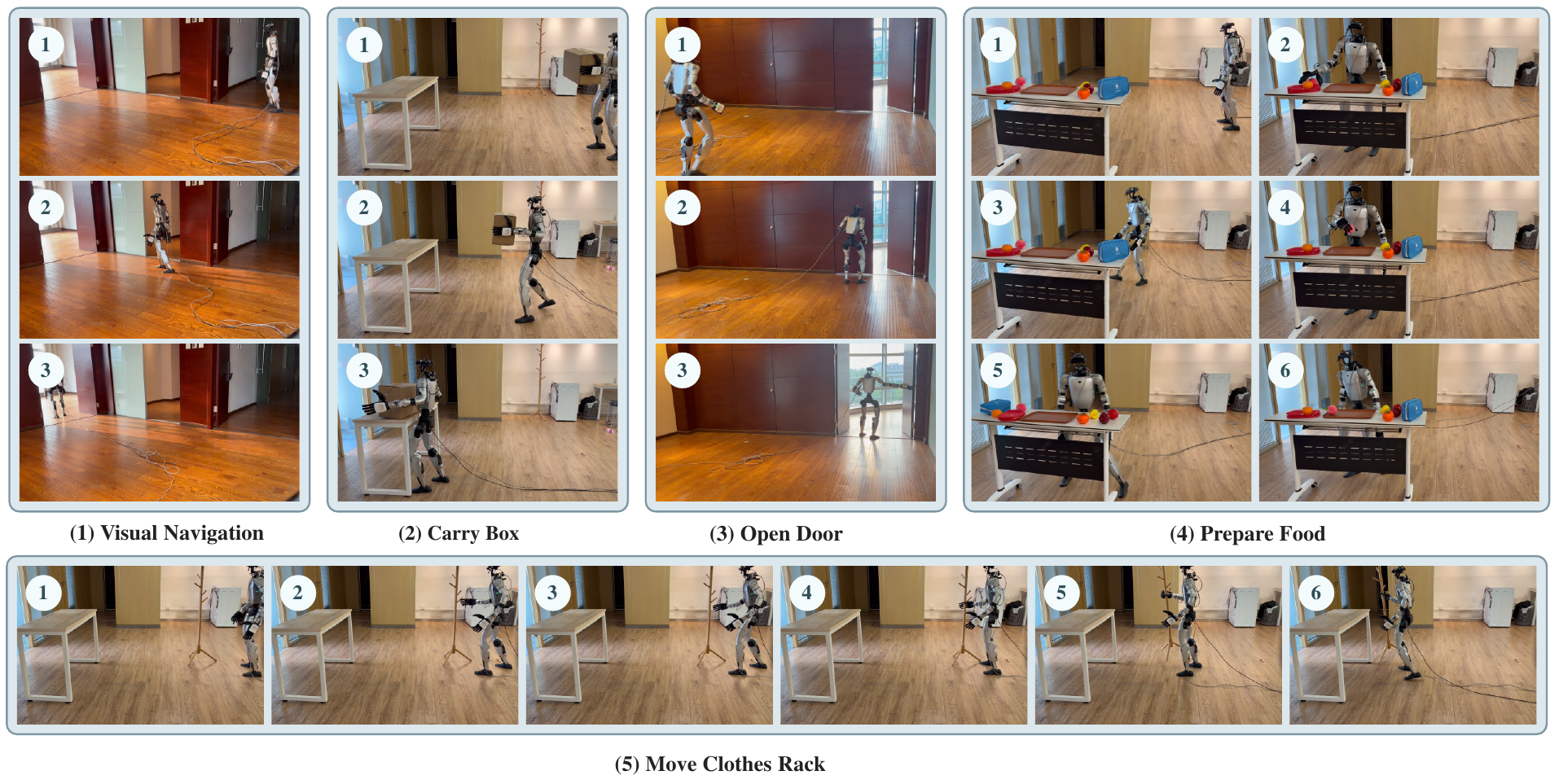}
    \caption{Real-world tasks: box transport, food preparation,
    visual navigation, clothes-rack manipulation, and door opening.}
    \label{fig:real_exp}
\end{figure}
\paragraph{Tasks and evaluation.}
The tasks in Figure~\ref{fig:real_exp} require different combinations
of locomotion and object interaction.
Box transport combines carrying with walking, while food preparation
involves several objects.
Visual navigation requires reaching a visually specified target.
Moving a clothes rack and opening a door require coordinated
body motion and hand contact.
The task instructions are fixed across methods:
\emph{Carry the box}, \emph{Prepare the food},
\emph{Navigate to the target}, \emph{Move the clothes rack},
and \emph{Open the door}.

Each method is trained with $200$ demonstrations per task,
for $1000$ demonstrations in total.
We evaluate each policy over $30$ trials per task under held-out
initial configurations, giving $150$ trials per method.
Trials vary box and rack positions, food objects and poses,
navigation targets, and doors and handles.
All methods use the same demonstrations, instructions, observations,
and evaluation scenes.
We report success rate for each task and the mean across the five tasks.

\begin{table}[t]
    \centering
    \small
    \caption{Real-world success rate (\%) over 30 trials per task.
    Each method is trained with 200 demonstrations per task.
    AVG is the unweighted mean across the five tasks.}
    \label{tab:real_world_results}
    \setlength{\tabcolsep}{2pt}
    \renewcommand{\arraystretch}{1.08}
    \begin{tabular*}{\linewidth}{@{\extracolsep{\fill}}lrrrrrr@{}}
        \toprule
        Method & \shortstack{Carry\\Box} & \shortstack{Prepare\\Food}
        & \shortstack{Visual\\Navigation} & \shortstack{Move\\Clothes Rack}
        & \shortstack{Open\\Door} & AVG \\
        \midrule
        GR00T~\citep{nvidia2025gr00tn1openfoundation} & 50.00 & 43.33 & 60.00 & 33.33 & 40.00 & 45.33 \\
        Fast-WAM~\citep{yuan2026fastwamworldactionmodels} & 53.33 & 50.00 & 56.67 & 43.33 & 46.67 & 50.00 \\
        w/o LF & 56.67 & 43.33 & 66.67 & 46.67 & 53.33 & 53.33 \\
        Ours & \textbf{73.33} & \textbf{70.00} & \textbf{76.67}
        & \textbf{66.67} & \textbf{70.00} & \textbf{71.33} \\
        \bottomrule
    \end{tabular*}
    \par\smallskip
    {\footnotesize Ours denotes \methodname{}; GR00T denotes GR00T N1.7.
    Trials vary box/rack position, food objects and poses, navigation
    targets, and doors and handles.\par}
\end{table}

\paragraph{Whole-body task completion.}
Language Forcing improves all five tasks and raises average SR
from $53.33\%$ to $71.33\%$
(Table~\ref{tab:real_world_results}).
Food preparation shows the largest gain, at $26.67$ percentage points.
The improvements extend from navigation, where progress means
reaching a target location, to manipulation, where completion
requires changing the state of an object.
The comparison with the same WAM shows that task-relevant semantic supervision complements action and future-frame prediction and improves real-world task success.

\paragraph{Deployment efficiency.}
With KV caching and quantization, policy inference takes
$91$\,ms on a RTX4090 24G GPU.
Language Forcing is used only during training, so our method does not add computation at inference.
More details are given in Appendix~\ref{sec:deployment_details}.

\section{Conclusion}
We introduced \methodname{}, a training strategy that uses language supervision to emphasize task-relevant changes in whole-body world--action models. Given current and future observations, a frozen VLM generates assessments of task progress and predictions of subsequent states. We align their language embeddings with intermediate future-visual representations, complementing the original visual and action objectives without adding inference overhead. Experiments on HumanoidArena and five real-world tasks demonstrate improved whole-body policy performance. Controlled studies further show the benefits of transition-specific language supervision and its effectiveness across training-data scales. These results support language supervision as a practical way to improve action learning from predictive visual representations.

% Keep the nine-page main paper separate from statements and references.
\clearpage
\endgroup

\subsection*{AI use statement}

We used generative AI tools for language polishing (e.g., grammar and clarity). The authors verified and edited all AI-suggested changes, and all technical content was produced and checked by the authors.
\subsection*{Ethics statement}
Our study does not involve human participants or user data, and we do not
collect, store, or release any personally identifiable information. We do not
anticipate any additional ethical concerns arising from the methods or
experiments presented in this paper.
\subsection*{Reproducibility statement}
To support reproducibility, we provide detailed training/evaluation settings in the appendix and include and videos in the supplementary materials, when permitted by the submission policy.
% \textcolor{yellow}{[DELETE/REPLACE: Template placeholder text. Replace with actual author contributions or remove this section for submission.]}

% \textcolor{yellow}{[DELETE/REPLACE: Template placeholder text. Replace with actual acknowledgments or remove this section for submission.]}

\bibliography{iclr2027_conference}
\bibliographystyle{iclr2027_conference}

\appendix
\section{Implementation Details}
\label{sec:implementation_details}

\subsection{Dataset and Evaluation Protocol}
\label{sec:data_protocol}

\paragraph{Demonstrations.}
We use the seven-task HumanoidArena benchmark
\citep{wang2026humanoidarenabenchmarkingegocentrichierarchical}.
For each task, we use the merged demonstration pool containing $100$ SONIC episodes and $100$ TWIST2 episodes.
The seven tasks therefore provide $1400$ successful episodes in total.
This is a mixed-GMT demonstration setting with deployment on SONIC.
The original recordings contain synchronized egocentric observations and whole-body state--action supervision at $50$\,Hz.
The benchmark's front-camera recordings have a resolution of $640\times480$; model-specific preprocessing converts these images to the input format required by each policy.
The retained trajectories contain $856{,}061$ frames and $788{,}861$ complete $48$-step training windows.
Table~\ref{tab:dataset_statistics} summarizes the retained demonstration pool.

\begin{table}[!htbp]
    \centering
    \caption{Demonstration statistics for the seven HumanoidArena tasks. Each task contains $100$ SONIC and $100$ TWIST2 episodes. A trajectory of length $L$ contributes $\max(0,L-48)$ complete windows.}
    \label{tab:dataset_statistics}
    \small
    \setlength{\tabcolsep}{6pt}
    \begin{tabular}{@{}llrrrr@{}}
        \toprule
        Suite & Task & SONIC & TWIST2 & Frames & Windows \\
        \midrule
        HOI & Football & 100 & 100 & 83,787 & 74,187 \\
        HOI & DoubleDesk & 100 & 100 & 165,898 & 156,298 \\
        HOI & P\&PBox & 100 & 100 & 115,053 & 105,453 \\
        HSI & OpenDoor & 100 & 100 & 130,310 & 120,710 \\
        HSI & SitSofa & 100 & 100 & 112,312 & 102,712 \\
        HSI & Boxing & 100 & 100 & 85,375 & 75,775 \\
        HSI & VisNavi & 100 & 100 & 163,326 & 153,726 \\
        \midrule
        \multicolumn{2}{l}{Total} & 700 & 700 & 856,061 & 788,861 \\
        \bottomrule
    \end{tabular}
\end{table}

\paragraph{Data allocation and evaluation.}
Our protocol does not require a separate demonstration validation set.
The Language-Forcing defaults are set from prior experience and
initial training-loss magnitudes, and every run uses the EMA checkpoint
at update $50000$.
Neither choice depends on evaluation success rates.
For the main comparison, all methods therefore use the same
$1400$ demonstration episodes for training.
Policy performance is measured through newly executed closed-loop
rollouts in simulation.
Table~\ref{tab:evaluation_protocol} summarizes the shared protocol.

\paragraph{Hyperparameters.}
For \methodname{} and its no-LF counterpart, we retain the post-training
optimization settings of GigaWorld-Policy 0.5.
The alignment layer and language-loss weight are specific to
Language Forcing and are set separately.
We choose block $22$ based on prior experience with supervision
of intermediate representations; block indices are zero-based.
We set $\lambda_{\mathrm{LF}}=1.0$ so that the weighted language loss
does not greatly exceed the action-flow loss at the start of training.
These defaults were determined before the ablation sweeps
and were not selected by comparing evaluation success rates.
We subsequently evaluated other layers and weights to examine
sensitivity to these choices.
The ablation results were not used to revise the default configuration.

\paragraph{Checkpoint selection.}
The humanoid state/action interface and the fixed $50$K-step
training budget are specified separately in this appendix.
We train each run to this budget and directly evaluate the EMA
weights at update $50000$.
Intermediate checkpoints are not ranked by evaluation success rate,
and we do not apply performance-based early stopping.
The LF and no-LF models therefore use the same base optimization
settings, training budget, and checkpoint-selection procedure.

\paragraph{Training seed.}
Given the cost of WAM training, repeating each configuration across
multiple training seeds is impractical within our available compute budget.
We use $42$, a commonly used training seed, for the matched LF
and no-LF runs and all WAM ablations.
The seed is fixed in advance and is not selected based on evaluation
success rates.

\paragraph{Online test construction.}
We evaluate four modes: base, visual, semantic, and execution.
Each task--mode pair uses group seeds $\{0,1,2\}$ with $20$ repetitions per seed, yielding $60$ rollouts per mode and $240$ rollouts per task.
All modes use SONIC as the execution backend, the benchmark's task-specific success criteria, and matched episode seeds and initial conditions across methods within each mode.
There are $7\times4\times3\times20=1680$ rollouts per method.
The $240$-rollout task score pools the four modes with equal weight; it is not a base-mode-only score.
Let $S_{k,m}$ be the number of successes for task $k$ and mode $m$.
We compute
\begin{equation}
    \mathrm{SR}_k=100\frac{\sum_{m=1}^{4}S_{k,m}}{240},
    \qquad
    \mathrm{AVG}=\frac{1}{7}\sum_{k=1}^{7}\mathrm{SR}_k.
\end{equation}
All counts are aggregated before rounding.

\begin{table}[!htbp]
    \centering
    \caption{Shared data and evaluation protocol. The aggregate task scores combine the base mode and three generalization modes with equal rollout counts.}
    \label{tab:evaluation_protocol}
    \small
    \renewcommand{\arraystretch}{1.10}
    \begin{tabular}{@{}p{0.32\linewidth}p{0.62\linewidth}@{}}
        \toprule
        Configuration & Setting \\
        \midrule
        Demonstration source & Merged SONIC and TWIST2 demonstrations \\
        Retained demonstration pool & 200 successful episodes per task; 1400 in total \\
        State / action dimension & 35 / 35, shared by all compared policies \\
        Evaluation modes & Base, visual, semantic, execution \\
        Group seeds / repetitions & $\{0,1,2\}$ / 20 per seed per mode \\
        Online test conditions & Matched episode seeds and initial robot states across methods within each mode \\
        Task success / termination & HumanoidArena task-specific criteria \\
        Deployment backend & SONIC for every evaluation mode \\
        Evaluation budget & 60 per task--mode; 240 per task; 1680 per method \\
        Aggregate metric & Unweighted mean SR over seven tasks \\
        Training budget & 50K gradient steps per method \\
        WAM checkpoint & EMA weights at update 50,000 \\
        \bottomrule
    \end{tabular}
\end{table}

\paragraph{Evaluation modes.}
The base mode uses the benchmark's original assets, object-pose ranges, and ambient illumination.
The execution mode expands object-pose randomization ranges while preserving the assets and lighting; for example, Football expands its ball-position range in both horizontal axes from $[-1.2,1.2]$ to $[-1.45,1.45]$.
The semantic mode substitutes benchmark-provided assets while preserving the task goal and affordance, including texture changes, distractor objects, and the Boxing target's sphere-to-cube substitution.
The visual mode replaces the ambient DomeLight with a directional DistantLight and randomizes pitch and roll over $[-20^\circ,20^\circ]$, intensity over $[3000,7000]$, and each RGB color component over $[0.65,0.85]$; yaw is fixed at zero.
Task-specific scene and reset settings follow the corresponding HumanoidArena SONIC evaluation configurations.
The simulator time step is $0.005$\,s with control decimation $4$, giving a $50$\,Hz control rate.

\begin{table}[!htbp]
    \centering
    \caption{Evaluation modes pooled in the main task-success scores. Each mode contributes $60$ rollouts per task.}
    \label{tab:evaluation_modes}
    \small
    \setlength{\tabcolsep}{4pt}
    \begin{tabular}{@{}llll@{}}
        \toprule
        Mode & Object-pose range & Assets & Lighting \\
        \midrule
        Base & Standard & Original & Ambient \\
        Execution & Expanded & Original & Ambient \\
        Semantic & Standard & Benchmark variants & Ambient \\
        Visual & Standard & Original & Randomized directional \\
        \bottomrule
    \end{tabular}
\end{table}

\paragraph{Temporal supervision and data subsets.}
For the WAM, each training window uses five visual frames at offsets $\{0,12,24,36,48\}$ and a $48$-step action trajectory.
At $50$\,Hz, the future endpoint is $0.96$\,s after the current observation.
For a trajectory of length $L$, valid anchors are $t=0,\ldots,L-49$.
Actions are taken at indices $t,\ldots,t+47$, while the final visual frame is at $t+48$.
Windows stay within one trajectory and are neither padded nor joined across episode boundaries.
The frozen teacher constructs the teacher-generated language target from the current image, the ground-truth endpoint, and the task context of that training window.
The data-efficiency study varies the fraction of eligible training trajectories over $\{10\%,40\%,70\%,100\%\}$.
For each fraction, the runs with and without Language-Forcing use the same trajectory subset.
Selection is performed on complete trajectories separately within each task: trajectories are ranked by a deterministic SHA-256 key using selection seed $42$, and the first $\lfloor fN_k\rfloor$ are retained for fraction $f$ and eligible task count $N_k$.
The ranking is fixed across fractions, producing nested subsets.
This sampling rule balances the retained fraction across tasks but does not enforce an additional equal-quota constraint for the two tracker sources.
All settings retain $50$K gradient updates and the same evaluation protocol.
These fractions refer to the training partition, not to validation trajectories or online evaluation rollouts.

\paragraph{Shuffled-language control.}
For the shuffled-language ablation, we reassign cached language targets across different trajectories within the same task, breaking their correspondence with the recipient window's temporal transition.
The reassignment is generated once with random seed $0$ and fixed in the language cache throughout training.
The observation/action windows, task-instruction conditioning, language-loss weight, and training schedule otherwise match the standard Language-Forcing run.
This control preserves task-level language context while disrupting transition-specific supervision.

\subsection{Baseline Implementations and Fair Comparison}
\label{sec:baseline_protocol}

\paragraph{Shared comparison conditions.}
All methods use the same selected demonstrations, task instructions, egocentric observation source, $35$D state and action interface, and SONIC evaluation backend.
All policies are trained for $50$K gradient steps.
We retain architecture-specific backbone initialization, image preprocessing, learning rate, batch size, precision, and temporal prediction design.
Thus, the comparison matches the task data, evaluation conditions, and gradient-step budget, while allowing the standard configurations of different policy families.

\paragraph{Baseline configurations.}
ACT, DP, FM, and $\pi_{0.5}$ are trained for $50{,}000$ updates with a batch size of $64$ and random seed $42$.
All four use one observation step, a $35$-dimensional state and action interface, an action chunk length of $20$, and $20$ inference action steps.
ACT, DP, and FM resize and pad the images to $224\times224$ and use float32; $\pi_{0.5}$ uses bfloat16.
ACT uses a ResNet-18 visual encoder and a learning rate of $1\times10^{-5}$.
DP also uses ResNet-18, with a learning rate of $1\times10^{-4}$ and a policy horizon of $24$.
FM uses CLIP ViT-B/16, a learning rate of $2\times10^{-5}$, and a policy horizon of $40$.
We initialize $\pi_{0.5}$ from its base checkpoint and use a learning rate of $2.5\times10^{-5}$.
We also fine-tune LingBot-VA, GR00T N1.7, Fast-WAM, and GigaWorld-Policy 0.5 on the same demonstration set for $50{,}000$ updates.
Each method uses the $35$-dimensional state and action interface and is evaluated in $240$ rollouts per task under the four-mode SONIC protocol.
\methodname{} and its no-LF counterpart share the GigaWorld-Policy 0.5 initialization, training data, preprocessing, horizons, and optimization settings.
\methodname{} adds the language projector and a language-alignment loss with weight $1.0$ at layer index $22$.
The no-LF model omits the projector and cached language targets.
Both retain the action-flow, visual-flow, and quaternion losses and use the same deployment path.

\subsection{Language-Forcing Training Configuration}
\label{sec:lf_training_configuration}

Table~\ref{tab:training_details} records the shared configuration used for the controlled Language-Forcing runs.
Subset experiments retain the same number of gradient updates and alter only the pool of eligible training trajectories.

\begin{table}[!htbp]
    \centering
    \caption{Training configuration for the controlled WAM experiments on HumanoidArena. Except for the Language-Forcing term and its auxiliary projector, the same configuration is used by the matched model without Language-Forcing.}
    \label{tab:training_details}
    \small
    \begin{tabularx}{\linewidth}{@{}>{\raggedright\arraybackslash}p{.54\linewidth}X@{}}
        \toprule
        Configuration & Value \\
        \midrule
        Initialization & GigaWorld-Policy 0.5 \\
        Input resolution (width $\times$ height) & $320\times384$ \\
        Visual frame offsets & $\{0,12,24,36,48\}$ \\
        Trajectory sampling rate & 50\,Hz \\
        Action horizon / dimension & $48$ / $35$ \\
        Teacher endpoint interval & $0.96$\,s \\
        Optimizer & CAME8Bit \\
        Learning rate & $6\times10^{-5}$ \\
        Learning-rate schedule / warmup & Constant / none \\
        Weight decay & $10^{-2}$ \\
        Global batch size & $128$ \\
        GPUs / per-GPU batch / accumulation & $8$ / $16$ / $1$ \\
        Gradient steps & $50$K \\
        Precision & bfloat16 \\
        Distributed optimizer & DeepSpeed ZeRO-2 \\
        Gradient clipping & $1.0$ \\
        Random seed & $42$ \\
        Action / visual flow shift & $5.0$ / $2.0$ \\
        Quaternion geometry / norm weight & $0.1$ / $0.01$ \\
        Default alignment layer index & $22$ (zero-based) \\
        Default language-loss weight & $1.0$ \\
        Language projector hidden / output dim. & $8192$ / $4096$ \\
        Evaluation weights & EMA at 50K steps \\
        \bottomrule
    \end{tabularx}
\end{table}

\subsection{State, Action, and Visual Preprocessing}
\label{sec:representation_details}

\paragraph{Unified whole-body interface.}
All compared policies use the same $35$D state and $35$D action representation.
The state contains $29$ joint positions, a four-dimensional root quaternion in Hamilton $(w,x,y,z)$ order, and two hand-state slots.
The hand-state slots are set to zero because the benchmark observation does not provide hand states.
We discard the joint velocities in the original $64$D benchmark observation.
Actions contain $29$ absolute reference joint positions, a root quaternion, and two native binary hand commands.
We convert the benchmark's continuous $6$D rotations to unit quaternions and reorder its interleaved joint coordinates into a common G1 joint order.
The $35$D action interface omits the root-translation and root-height fields of the original $40$D action representation.
All policies use this same adapter for comparison; the LF and no-LF WAM additionally share the same model architecture and initialization.

\paragraph{Normalization and quaternion continuity.}
For joint $j$, let $q_j^{01}$ and $q_j^{99}$ denote the first and ninety-ninth percentiles of its position in the normalization statistics.
These statistics are recomputed from the final seven-task demonstration data and shared by the state and action transforms.
We apply the same affine transform to state and action joint positions,
\begin{equation}
    \widetilde{x}_j = 2\frac{x_j-q_j^{01}}{q_j^{99}-q_j^{01}}-1,
\end{equation}
without clipping values outside $[-1,1]$.
Quaternions are unit-normalized and hand targets retain their $0/1$ values.
Within each action chunk, the sign of each target quaternion is chosen to have a nonnegative dot product with the preceding quaternion, starting from the current root orientation.
This removes sign discontinuities between equivalent quaternion representations.

\paragraph{Visual preprocessing and latent caching.}
The WAM uses RGB images of size $384\times320$ (height $\times$ width),
with pixel values normalized to $[-1,1]$.
The frozen Wan2.2-TI2V-5B visual autoencoder encodes each five-frame
clip and the current frame separately.
We cache the normalized latents in bfloat16 and reuse them during training.

\subsection{Offline Language-Target Construction}
\label{sec:language_target_details}

\paragraph{VLM inputs.}
We use the frozen instruction-tuned Gemma-4-12B model to construct
language targets.
The current frame $I_t$ and future frame $I_{t+48}$ are placed side
by side, with the current frame on the left.
The VLM also receives the task context and elapsed time ($0.96$ seconds)
to assess task progress and predict future states.

\paragraph{Prompt pool and offline sampling.}
For each task, we construct 20 semantically equivalent prompt pairs in a single generation request.
Each pair contains a system message followed by a task-specific user message; the wording of both messages may vary while preserving the task and response format.
During offline cache construction, we sample one pair for each training window and instantiate its task context and elapsed time before running the teacher.
The selected prompt and resulting language target are fixed for that window and reused across training epochs.
Appendix~\ref{app:language-prompts} presents the prompt templates and task-wise examples.

\paragraph{Response format.}
The teacher returns two English paragraphs, headed task progress and future prediction, in that order.
The first describes the robot's visible state and actions, spatial relationships, and relevant colors, textures, and apparent materials.
The second describes a possible near-term continuation beyond the observed endpoint.
Both input panels are observations; the prediction concerns subsequent behavior.
The following Football example is one representative system/user pair from the prompt pool.

\begin{lfappbpromptbox}[breakable=false]\small
\textbf{System prompt}\par
``Describe the humanoid's visible state, actions, spatial relations, and relevant colors and apparent materials. Write two paragraphs headed Task progress: and Future prediction:, in that order. Keep predictions tentative and grounded in observed behavior; describe the robot without referring to images or frames.''
\medskip\par
\textbf{User prompt}\par
``Task context: Kick the football into the goal.\\
Elapsed time from CURRENT to FUTURE: 0.96 seconds.\\
Describe the robot's progress toward kicking the football into the goal and its likely next movement.''
\end{lfappbpromptbox}

\paragraph{Target encoding and caching.}
We encode the complete two-paragraph teacher response with the frozen UMT5 encoder and tokenizer distributed with Wan2.2-TI2V-5B.
The target tokenizer uses a maximum length of $256$ tokens with truncation and batch padding.
We take the mean of the final hidden states over non-padding tokens, producing one $4096$D vector, and store it in bfloat16.
Each response and embedding is keyed by episode index and window start frame.
Cache checks validate key uniqueness, window coverage, source timestamps, prompt version, and finite embeddings.
Prompt sampling, Gemma inference, and target encoding are completed before policy optimization.
The training loop loads the cached target embeddings and reuses them across epochs.
The target branch is separate from task-instruction conditioning: task text is cached as up to $64$ token-level UMT5 features, whereas the teacher-generated language target is a pooled vector used only for training.

\subsection{Architecture and Training Objectives}
\label{sec:architecture_objectives}

\paragraph{Backbone and adaptation.}
We initialize the transformer from GigaWorld-Policy 0.5 and use its $30$-block mixture-of-transformers backbone.
The visual expert has hidden width $3072$ and feed-forward width $14336$; the action expert has hidden width $1024$ and feed-forward width $4096$.
Attention uses $24$ heads, and visual latents are patchified with a $1\times2\times2$ kernel.
The current observation contributes $120$ reference tokens and the future latent contributes $120$ future tokens; the state and action streams contain one and $48$ tokens, respectively.
For the humanoid embodiment, we add the state/action input and action output projections for the $35$D interface while loading the compatible pretrained backbone parameters.
The original embodiment-specific projection rows are preserved; the new humanoid projections and backbone are trained.
Gemma, UMT5, and the visual autoencoder remain frozen.

\paragraph{Attention and alignment.}
State and reference queries attend to the state and reference tokens.
Action queries additionally attend to action tokens, while future-visual queries can attend to all four token groups.
Task-instruction features condition the streams through cross-attention.
We take the future-visual hidden states after the selected transformer block and exclude all current-reference tokens from pooling.
Layer indices are zero-based throughout the alignment sweep; index $22$ denotes the output of the twenty-third block.
The language projector is LayerNorm$(3072)$, Linear$(3072,8192)$, GELU, and Linear$(8192,4096)$.
The pooled future representation remains connected to the computation graph, so its cosine-alignment loss updates the predictor and the action-conditioned representations that it reads.

\paragraph{Flow matching.}
We sample action and visual noise levels independently.
For either stream, draw $u\sim\mathcal{U}(0,1)$ and apply
\begin{equation}
    \sigma_s(u)=\frac{s u}{1+(s-1)u},
\end{equation}
with $s=5$ for actions and $s=2$ for visual latents.
We discretize this value to the nearest multiple of $1/1000$.
For a clean target $x$ and independent Gaussian noise $\epsilon$, the noisy input is
$x_\sigma=(1-\sigma)x+\sigma\epsilon$, and the flow target is $\epsilon-x$.
The current observation and robot state remain clean.
The action loss averages squared flow error over all $48\times35$ entries.
For the visual loss, the reference-frame slice is masked out, and the squared masked residual is averaged over the full clip-latent tensor.
The action-flow and visual-flow terms each have coefficient $1$.

\paragraph{Quaternion losses and total objective.}
We reconstruct a clean action estimate as $\widehat{x}=x_\sigma-\sigma v_\theta(x_\sigma)$.
Let $\widehat{q}$ be its four quaternion coordinates, $\overline{q}=\widehat{q}/\|\widehat{q}\|_2$, and $q$ the unit-normalized target.
Averaged over the action chunk and batch, the additional losses are
\begin{align}
    \mathcal{L}_{\mathrm{quat}} &= \mathbb{E}\left[1-|\langle\overline{q},q\rangle|\right],\\
    \mathcal{L}_{\mathrm{norm}} &= \mathbb{E}\left[(\|\widehat{q}\|_2-1)^2\right].
\end{align}
Quaternion normalization uses an $\epsilon$ floor of $10^{-8}$.
The complete training objective is
\begin{equation}
    \mathcal{L}=\mathcal{L}_{\mathrm{action}}+\mathcal{L}_{\mathrm{visual}}
    +0.1\mathcal{L}_{\mathrm{quat}}+0.01\mathcal{L}_{\mathrm{norm}}
    +\lambda_{\mathrm{LF}}\mathcal{L}_{\mathrm{LF}}.
\end{equation}
The default language-loss weight is $1.0$.
The no-LF implementation omits the language projector and language-target cache, while retaining task-text conditioning and all other losses.

\paragraph{Optimization.}
We use the CAME8Bit optimizer.
Its three momentum coefficients are $0.9$, $0.999$, and $0.9999$.
The numerical-stability constants are $10^{-30}$ and $10^{-16}$; the internal update clipping threshold is $1.0$.
Optimizer-state quantization uses blocks of $2048$ elements and the default eligibility threshold of $16384$ elements.
The learning rate is constant at $6\times10^{-5}$ without warmup, weight decay is $10^{-2}$, and the global gradient-norm limit is $1.0$.
Training uses eight GPUs with $16$ windows per GPU, one gradient-accumulation step, bfloat16 precision, and DeepSpeed ZeRO-2.
The training seed is $42$.
The window sampler shuffles all eligible windows, using its separate base seed $6666$ with an epoch offset; it does not impose an equal task-sampling quota.
Checkpoints are saved every $5000$ updates and evaluation uses the EMA transformer at update $50000$.
EMA is updated each optimization step with a decay that increases according to the implementation's $(1+s)/(10+s)$ schedule, capped at $0.9999$, where $s$ is the offset optimization step; the first update copies the current parameters.

\subsection{Closed-Loop Deployment}
\label{sec:deployment_details}

The policy encodes the current image and conditions on the $35$D state and cached task-instruction features.
It generates a $48\times35$ action chunk from Gaussian noise using $10$ Euler flow-matching steps with shift $5$ and without classifier-free guidance.
We denormalize the joint targets, convert them back to SONIC's joint order, unit-normalize predicted root quaternions, and clamp hand outputs to $[0,1]$ before passing them to the tracker interface.
The controller executes $20$ actions at $50$\,Hz before requesting a new chunk, corresponding to a nominal replanning interval of $0.4$\,s in control time.
The complete predicted chunk remains available to SONIC's future-reference window.
The adapter converts episode-relative root orientation to the world-frame reference using the initial heading and obtains joint-velocity references from consecutive joint positions at $50$\,Hz.
The final deployment implementation omits Gemma, the teacher-response encoder and language-target cache, the language projector, and future-visual rollout tensors.
The current-image encoder and cached task-instruction conditioning remain part of the policy.

\paragraph{Inference acceleration and latency.}
The deployment implementation uses KV caching and quantized inference.
Within each action-generation call, the observation context is fixed while the action sample is updated over the $10$ Euler steps.
We compute the context attention keys and values once and reuse them across these steps, avoiding repeated context computation.
The observation-dependent cache is refreshed when a new observation is received for replanning.
The deployed policy has a mean input-to-action latency of $91$\,ms on an NVIDIA RTX 4090 (24\,GB), measured from receiving the policy inputs to returning the predicted action chunk.
Table~\ref{tab:deployment_efficiency} summarizes the inference settings.

\begin{table}[!htbp]
    \centering
    \caption{Policy inference and execution settings. Policy inference latency and reference-command frequency describe different stages of deployment.}
    \label{tab:deployment_efficiency}
    \small
    \begin{tabularx}{\linewidth}{@{}>{\raggedright\arraybackslash}p{.46\linewidth}X@{}}
        \toprule
        Configuration & Value \\
        \midrule
        Deployment GPU & NVIDIA RTX 4090 (24\,GB) \\
        Inference acceleration & KV caching and quantization \\
        Action chunk & $48\times35$ \\
        Action sampler & Euler, $10$ steps, shift $5$ \\
        Classifier-free guidance & Disabled \\
        Mean input-to-action latency & $91$\,ms \\
        Reference-command frequency & $50$\,Hz \\
        Executed actions before replanning & $20$ \\
        Nominal replanning interval & $0.4$\,s in control time \\
        \bottomrule
    \end{tabularx}
\end{table}

\paragraph{Episode termination.}
Evaluation uses the benchmark task-specific success criteria and treats falls and timeouts as failures.
Maximum control steps for Football, DoubleDesk, P\&PBox, OpenDoor, SitSofa, Boxing, and VisNavi are $1300$, $2000$, $2000$, $1800$, $1050$, $1500$, and $1800$, respectively.
Infrastructure failures are retried rather than being recorded as task outcomes.
For reproducible initializations, the benchmark derives each episode seed from the UTF-8 string formed by the task identifier, group seed, and repeat index, separated by vertical bars.
It takes the final four bytes of the SHA-256 digest as an unsigned big-endian integer and masks the result to $31$ bits.

\paragraph{Software.}
The training implementation uses Python $3.11$, PyTorch $2.8.0$, torchvision $0.23.0$, Diffusers $0.36.0$, Accelerate $1.13.0$, DeepSpeed $0.17.4$, and LeRobot $0.4.4$, with the PyTorch CUDA $12.8$ build.
Teacher generation runs in a separate environment supporting Gemma-4; it is not executed during policy training or inference.

\subsection{Real-World Data and Deployment}
\label{sec:real_world_implementation}

\paragraph{Training data and initialization.}
We first train on the seven-task simulation demonstration set without adding real-robot data or demonstrations from other simulation sources.
For real-world deployment, we continue training from the simulation-trained checkpoint using our collected demonstrations.
We collect $200$ demonstrations for each of the five tasks, and train a single policy jointly across all five tasks.
All compared methods use the same real-world training data, task instructions, visual observations, and evaluation scenes.
Each method retains its simulation training configuration during real-world training, including the $50$K-update budget, batch size, learning rate, and optimization settings.
For \methodname{} and its no-LF counterpart, this uses CAME8Bit with a learning rate of $6\times10^{-5}$, a global batch size of $128$, and the final $50$K-step EMA checkpoint.

\paragraph{Evaluation and success criteria.}
We evaluate each task in $30$ trials under held-out initial configurations, totaling $150$ trials per method.
A trial is successful only when the full task is completed; partial completion receives no credit.
The same success criteria are used for all methods:

\begin{itemize}
    \setlength{\itemsep}{2pt}
    \setlength{\parsep}{0pt}
    \item \textbf{Carry the box.} The robot picks up the box, carries it to the designated table, and places it on the tabletop. The box remains supported by the table after the robot releases it.
    \item \textbf{Prepare the food.} The robot selects the specified ingredients and sorts each into its designated container or location. All requested ingredients must be correctly placed; missing an ingredient or placing it in the wrong destination counts as failure.
    \item \textbf{Navigate to the target.} The robot reaches the designated destination and comes to a stable stop without falling.
    \item \textbf{Move the clothes rack.} The robot moves the rack into the designated destination area. The rack remains upright and stationary when the robot stops moving it.
    \item \textbf{Open the door.} The robot operates the handle and opens the door sufficiently to pass through the doorway. Contacting the handle or moving the door without creating a traversable opening does not count as success.
\end{itemize}

Table~\ref{tab:real_world_data} summarizes the collected task demonstrations, evaluation settings, and hardware.

\begin{table}[!htbp]
    \centering
    \caption{Collected real-world task demonstrations, evaluation protocol, and hardware. Task names correspond to Section~\ref{sec:real_world_experiments}.}
    \label{tab:real_world_data}
    \label{tab:real_world_hardware}
    \small
    \setlength{\tabcolsep}{5pt}
    \renewcommand{\arraystretch}{1.10}
    \textbf{Task data and evaluation}\par\smallskip
    \begin{tabular}{@{}llrr@{}}
        \toprule
        Task & Evaluation variation & Collected demos & Trials \\
        \midrule
        Carry the box & Position & 200 & 30 \\
        Prepare the food & Object + pose & 200 & 30 \\
        Navigate to the target & Target & 200 & 30 \\
        Move the clothes rack & Position & 200 & 30 \\
        Open the door & Door + handle & 200 & 30 \\
        \midrule
        Total & -- & 1000 & 150 \\
        \bottomrule
    \end{tabular}
    \par\medskip
    \textbf{Hardware}\par\smallskip
    \begin{tabular}{@{}p{0.31\linewidth}p{0.63\linewidth}@{}}
        \toprule
        Component & Specification \\
        \midrule
        Humanoid platform & Unitree G1 (29-DoF) \\
        RGB-D camera & Orbbec Gemini 336L \\
        Demonstration capture & PICO 4 Ultra \\
        Deployment workstation & Local GPU workstation with NVIDIA RTX 4090 (24\,GB) \\
        \bottomrule
    \end{tabular}
\end{table}

% Keep Appendix A tables before Appendix B.
\FloatBarrier
\begingroup
\section{Language prompts and qualitative examples}
\label{app:language-prompts}
\subsection{Prompt templates}
\label{app:prompt-templates}
Each prompt consists of a system message followed by a user message. We show 20 system-message variants and 20 task-specific user requests for Football and OpenDoor. The corresponding entries form 20 prompt pairs for each task. A pair is selected for each training window during offline cache construction. The visual input contains two observed endpoints, CURRENT and FUTURE; \textit{Future prediction} refers to a possible continuation beyond the observed endpoint.

\begin{lfappbpromptbox}\small
\textbf{SYSTEM PROMPT TEMPLATES}\par\medskip

``Describe the humanoid's visible state, actions, spatial relations, and relevant colors and apparent materials. Write two paragraphs headed Task progress: and Future prediction:, in that order. Keep predictions tentative and grounded in observed behavior; describe the robot without referring to images or frames.''\par\vspace{1pt}
``Give a grounded account of humanoid behavior, including posture, object interaction, and visible appearance details. Use Task progress: for the observed task state, followed by Future prediction: for a plausible next action. Write two paragraphs, avoid image references, and qualify uncertain claims.''\par\vspace{1pt}
``Explain what the robot is visibly doing and how its behavior may continue. Include relevant object colors, textures, and apparent materials. Start the two paragraphs with Task progress: and Future prediction:, respectively. Focus on the robot and distinguish observed behavior from uncertain prediction.''\par\vspace{1pt}
``Describe task progress through the humanoid's pose, movement, and relationship to nearby objects. Mention useful color and material cues. Then provide a cautious near-term continuation. Use exactly two paragraphs, headed Task progress: and Future prediction:, without narrating individual images or frames.''\par\vspace{1pt}
``Report the robot's visible task state and likely next movement in two paragraphs: Task progress: followed by Future prediction:. Include posture, spatial relationships, and relevant object appearance or apparent material. Keep the account grounded in visual evidence and avoid referring to the images themselves.''\par\vspace{1pt}
``Summarize the humanoid's observed behavior with concrete details about its body configuration, object relationships, colors, and apparent materials. Follow with a qualified prediction. Label the two paragraphs Task progress: and Future prediction:, in order, and describe robot behavior without mentioning frames.''\par\vspace{1pt}
``Analyze the robot's visible progress and a plausible immediate continuation. Include relevant posture, object placement, color, texture, and apparent material. Respond with Task progress: first and Future prediction: second, each as a paragraph. Avoid image-by-image narration and unsupported certainty.''\par\vspace{1pt}
``Write a detailed account of the humanoid's task state, followed by a cautious forecast of its next action. Use the paragraph headings Task progress: and Future prediction:. Ground the description in visible body and object states, include appearance details, and avoid image references.''\par\vspace{1pt}
``Describe the robot's current behavior, its relation to task objects, and their relevant colors and apparent materials. Give a plausible next step separately. Use two paragraphs beginning Task progress: and Future prediction:. Refer directly to the robot, and mark uncertainty in interpretation or prediction.''\par\vspace{1pt}
``Explain the visible stage of the humanoid task using posture, movement, spatial relations, and relevant appearance details such as color and apparent material. Return Task progress: and then Future prediction: as two paragraphs. Keep the forecast tentative and omit references to individual frames.''\par\vspace{1pt}
``Provide a grounded description of the robot's task progress and a qualified short-term prediction. Include concrete pose and object details, with visible colors, textures, and apparent materials where relevant. Use two paragraphs headed Task progress: and Future prediction:, without describing the images as images.''\par\vspace{1pt}
``Characterize what the humanoid is doing and what it may do next. Describe relevant body configuration, object relationships, colors, and apparent materials. Present the observed state under Task progress: and the tentative continuation under Future prediction:. Use separate paragraphs and avoid frame references.''\par\vspace{1pt}
``Report visible robot behavior with useful details about stance, limb placement, nearby objects, and their color or apparent material. Then anticipate a plausible continuation. Format the response as Task progress: followed by Future prediction: in two paragraphs, keeping uncertain claims qualified and avoiding image narration.''\par\vspace{1pt}
``Describe the humanoid's task-relevant state and movement in concrete visual terms, including relevant colors, textures, and apparent materials. Follow with a cautious next-action prediction. Write two paragraphs beginning Task progress: and Future prediction:, respectively; discuss robot behavior without naming images or frames.''\par\vspace{1pt}
``Explain the robot's progress from its visible posture and interactions, using object appearance and spatial details when informative. Include colors and apparent materials. Then give a qualified near-term forecast. Use Task progress: and Future prediction: as paragraph headings and avoid references to individual images.''\par\vspace{1pt}
``Give an evidence-grounded account of the humanoid's current action and a plausible next movement. Include relevant body pose, object location, color, texture, and apparent material. Separate the answer into Task progress: and Future prediction: paragraphs; avoid frame narration and distinguish prediction from observation.''\par\vspace{1pt}
``Describe the robot's task state with attention to limb configuration, object relationships, and visible color or material cues. Predict a possible immediate continuation without assuming success. Use two paragraphs, Task progress: then Future prediction:, and refer to the robot rather than the images.''\par\vspace{1pt}
``Summarize the humanoid's visible progress and how its action might continue. Supply useful details about posture, spatial relationships, colors, and apparent materials. Begin the two paragraphs with Task progress: and Future prediction:. Keep uncertain interactions and predictions qualified, and do not mention frames.''\par\vspace{1pt}
``Describe what is visibly happening around the robot and how it relates to the task, including relevant posture, object appearance, color, and apparent material. Follow with a tentative next step. Use Task progress: and Future prediction: in separate paragraphs without image-by-image narration.''\par\vspace{1pt}
``Provide a concrete description of the humanoid's observed state and a cautious prediction of its next task-relevant action. Include spatial relationships and useful color, texture, or apparent-material details. Write Task progress: before Future prediction: in two paragraphs, without referring to images or frames.''\par\vspace{1pt}
\end{lfappbpromptbox}

The user message has the following form, with the task context and elapsed time instantiated for the selected observation pair. Only the final request varies across the quoted task-specific candidates below.
\begin{lfappbpromptbox}\small
\textbf{User prompt}\par
Task context: \texttt{\{task\}}\\
Elapsed time from CURRENT to FUTURE: \texttt{\{elapsed\_time\}}.\\
\texttt{\{sampled request\}}
\end{lfappbpromptbox}

\begin{lfappbpromptbox}\small
\textbf{FOOTBALL --- USER PROMPT TEMPLATES}\par\medskip
``Describe the robot's progress toward kicking the football into the goal and its likely next movement.''\par\vspace{1pt}
``Explain the robot's current stance relative to the football and goal, then predict a plausible next step.''\par\vspace{1pt}
``Summarize the visible football-task progress and describe how the robot may continue toward the goal.''\par\vspace{1pt}
``Report the robot's approach to or interaction with the football and its likely immediate continuation.''\par\vspace{1pt}
``Describe the robot's visible posture and ball-relative position, followed by its probable next action.''\par\vspace{1pt}
``Explain how the robot is positioned to advance the football task and what it may do next.''\par\vspace{1pt}
``Characterize the robot's current interaction with the football and predict the next short movement.''\par\vspace{1pt}
``Describe the visible state of the robot, football, and goal, then give a near-term prediction.''\par\vspace{1pt}
``Summarize task progress through the robot's posture and ball relationship, then anticipate the next action.''\par\vspace{1pt}
``Explain the robot's current football behavior and the most plausible immediate continuation.''\par\vspace{1pt}
``Report visible progress in approaching or kicking the ball, then describe what may happen next.''\par\vspace{1pt}
``Describe the robot's position and limb configuration around the football and its likely next adjustment.''\par\vspace{1pt}
``Give a grounded account of the robot's football-task state and a cautious short-term prediction.''\par\vspace{1pt}
``Describe how the robot is engaging with the football and how this behavior may continue.''\par\vspace{1pt}
``Summarize the visible robot-ball interaction, then predict a plausible continuation toward the task goal.''\par\vspace{1pt}
``Explain the current stage of the football task using visible robot state, then anticipate the next movement.''\par\vspace{1pt}
``Describe the robot's stance, the football's relative location, and the likely next task-relevant action.''\par\vspace{1pt}
``Report what the robot is visibly doing in the football task and what it is likely to do next.''\par\vspace{1pt}
``Characterize the visible task state around the robot and football, followed by a near-term forecast.''\par\vspace{1pt}
``Describe current progress toward the football goal and a plausible next step grounded in the robot's state.''\par\vspace{1pt}
\end{lfappbpromptbox}
\begin{lfappbpromptbox}\small
\textbf{OPENDOOR --- USER PROMPT TEMPLATES}\par\medskip
``Describe the robot's progress in opening the door and its likely next movement.''\par\vspace{1pt}
``Explain the robot's current position relative to the door and handle, then predict a plausible next step.''\par\vspace{1pt}
``Summarize the visible door-opening progress and describe how the robot may continue.''\par\vspace{1pt}
``Report the robot's approach to or interaction with the door handle and its likely immediate continuation.''\par\vspace{1pt}
``Describe the robot's visible posture and hand-handle relationship, followed by its probable next action.''\par\vspace{1pt}
``Explain how the robot is positioned to advance the door-opening task and what it may do next.''\par\vspace{1pt}
``Characterize the robot's current interaction with the door and predict the next short movement.''\par\vspace{1pt}
``Describe the visible robot, handle, and door configuration, then give a near-term prediction.''\par\vspace{1pt}
``Summarize task progress through the robot's posture and door relationship, then anticipate the next action.''\par\vspace{1pt}
``Explain the robot's current door-opening behavior and the most plausible immediate continuation.''\par\vspace{1pt}
``Report visible progress in approaching or opening the door, then describe what may happen next.''\par\vspace{1pt}
``Describe the robot's position and arm configuration near the door and its likely next adjustment.''\par\vspace{1pt}
``Give a grounded account of the robot's door-opening state and a cautious short-term prediction.''\par\vspace{1pt}
``Describe how the robot is engaging with the door handle and how this behavior may continue.''\par\vspace{1pt}
``Summarize the visible robot-door interaction, then predict a plausible continuation toward the task goal.''\par\vspace{1pt}
``Explain the current stage of door opening using visible robot state, then anticipate the next movement.''\par\vspace{1pt}
``Describe the robot's stance, hand placement, and the door opening, then predict the next relevant action.''\par\vspace{1pt}
``Report what the robot is visibly doing at the door and what it is likely to do next.''\par\vspace{1pt}
``Characterize the visible task state around the robot and door, followed by a near-term forecast.''\par\vspace{1pt}
``Describe current progress toward opening the door and a plausible next step grounded in the robot's state.''\par\vspace{1pt}
\end{lfappbpromptbox}

\subsection{Prompt generation and offline sampling}
\label{app:prompt-sampling}
For each task, we construct a pool of 20 semantically equivalent prompt pairs in a single generation request. Each pair contains a system message followed by a task-specific user message. Both messages may vary in wording while preserving the task and the two-paragraph response format. During offline cache construction, we sample one pair from the task-specific pool for each training window, instantiate the task context and elapsed time, and provide the paired observations to the frozen Gemma teacher. The resulting response is encoded with the frozen UMT5 encoder and stored with the corresponding window. Prompt sampling, teacher inference, and target encoding are completed before policy optimization. Training reuses the cached targets across epochs, avoiding repeated teacher computation.

\clearpage
\subsection{Task-wise input--response examples}
\label{app:qualitative-language}
Each constructed example presents the image input, prompt input, and two-paragraph output. CURRENT and FUTURE are observed endpoints shown in chronological order; the elapsed time is measured from the recording.

\noindent\textbf{Football}\par\vspace{3pt}
\lfappbexamplecard{}{\includegraphics[width=\linewidth,height=1.0in,keepaspectratio]{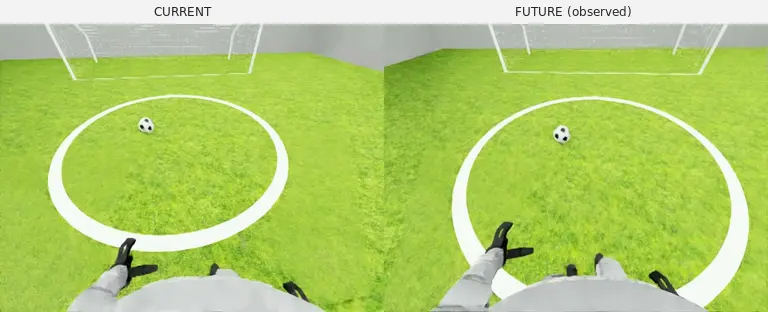}}{\textbf{System prompt}\newline ``Provide a grounded description of the robot's task progress and a qualified short-term prediction. Include concrete pose and object details, with visible colors, textures, and apparent materials where relevant. Use two paragraphs headed Task progress: and Future prediction:, without describing the images as images.''\par\vspace{2pt}\textbf{User prompt}\newline ``Task context: Kick the football into the goal.\newline Elapsed time from CURRENT to FUTURE: 0.967 seconds (recording time).\newline Report visible progress in approaching or kicking the ball, then describe what may happen next.''}{The robot is approaching a black-and-white football on bright green turf, with its forearms held forward and the white goal ahead. The ball remains several steps away, and the visible posture is consistent with closing the distance rather than making contact.}{The robot may continue forward and adjust its heading to bring the ball within kicking range while keeping the goal in front.}{}
\par\vspace{5pt}
\lfappbexamplecard{}{\includegraphics[width=\linewidth,height=1.0in,keepaspectratio]{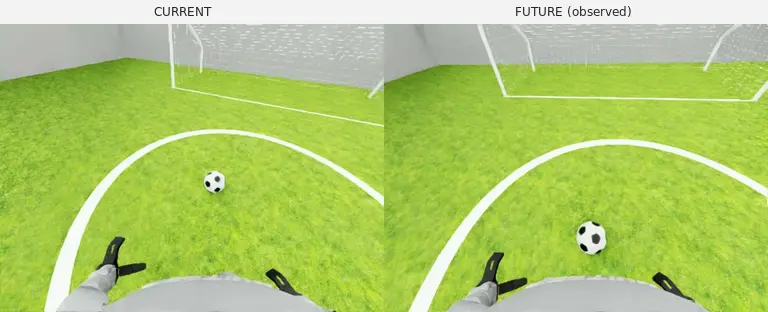}}{\textbf{System prompt}\newline ``Report the robot's visible task state and likely next movement in two paragraphs: Task progress: followed by Future prediction:. Include posture, spatial relationships, and relevant object appearance or apparent material. Keep the account grounded in visual evidence and avoid referring to the images themselves.''\par\vspace{2pt}\textbf{User prompt}\newline ``Task context: Kick the football into the goal.\newline Elapsed time from CURRENT to FUTURE: 0.967 seconds (recording time).\newline Describe the robot's visible posture and ball-relative position, followed by its probable next action.''}{The robot is closer to the black-and-white football near a curved white field marking. Its arms remain out in front, while the ball occupies a lower, nearer position relative to the body. No clear foot-ball contact is visible.}{The robot is likely to make a short positioning adjustment before bringing a foot toward the ball for a kick.}{}
\par\vspace{5pt}
\clearpage
\noindent\textbf{Football}\par\vspace{3pt}
\lfappbexamplecard{}{\includegraphics[width=\linewidth,height=1.35in,keepaspectratio]{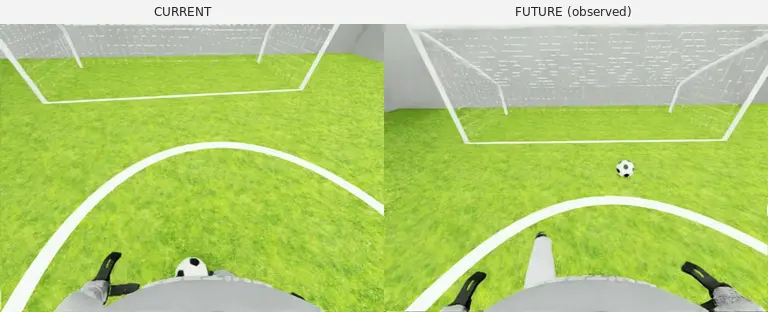}}{\textbf{System prompt}\newline ``Describe the robot's task state with attention to limb configuration, object relationships, and visible color or material cues. Predict a possible immediate continuation without assuming success. Use two paragraphs, Task progress: then Future prediction:, and refer to the robot rather than the images.''\par\vspace{2pt}\textbf{User prompt}\newline ``Task context: Kick the football into the goal.\newline Elapsed time from CURRENT to FUTURE: 0.967 seconds (recording time).\newline Describe the robot's stance, the football's relative location, and the likely next task-relevant action.''}{The robot extends a leg toward the black-and-white football on the green turf. The ball has moved farther ahead toward the white goal, and the extended leg is consistent with a kicking motion; the exact contact instant is not resolved.}{The robot may lower its extended leg and regain a stable stance while the ball continues toward the goal.}{}
\par\vspace{5pt}
\lfappbexamplecard{}{\includegraphics[width=\linewidth,height=1.35in,keepaspectratio]{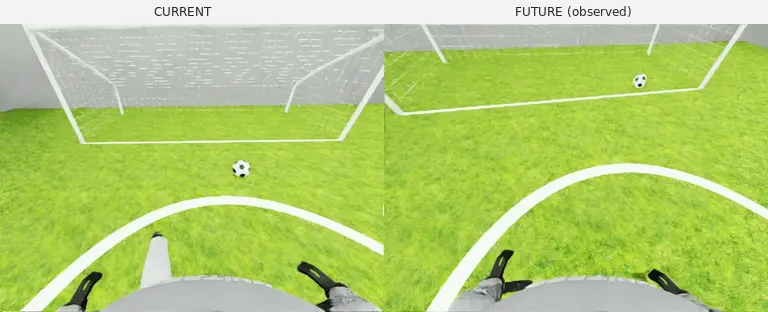}}{\textbf{System prompt}\newline ``Describe task progress through the humanoid's pose, movement, and relationship to nearby objects. Mention useful color and material cues. Then provide a cautious near-term continuation. Use exactly two paragraphs, headed Task progress: and Future prediction:, without narrating individual images or frames.''\par\vspace{2pt}\textbf{User prompt}\newline ``Task context: Kick the football into the goal.\newline Elapsed time from CURRENT to FUTURE: 0.967 seconds (recording time).\newline Report the robot's approach to or interaction with the football and its likely immediate continuation.''}{The black-and-white football is farther from the robot, near the white goal area on the green field. The previously extended leg is no longer prominent, and both forearms remain ahead of the torso as the robot returns toward a ready posture.}{The robot may stabilize its stance and observe the ball near the goal; the final scoring outcome is not established by this view.}{}
\par\vspace{5pt}
\clearpage
\noindent\textbf{DoubleDesk}\par\vspace{3pt}
\lfappbexamplecard{}{\includegraphics[width=\linewidth,height=1.35in,keepaspectratio]{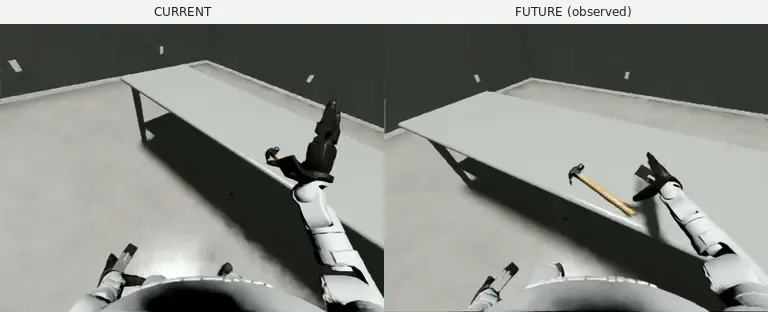}}{\textbf{System prompt}\newline ``Describe the humanoid's visible state, actions, spatial relations, and relevant colors and apparent materials. Write two paragraphs headed Task progress: and Future prediction:, in that order. Keep predictions tentative and grounded in observed behavior; describe the robot without referring to images or frames.''\par\vspace{2pt}\textbf{User prompt}\newline ``Task context: Move the hammer from the source desk to the basket on the other desk.\newline Elapsed time from CURRENT to FUTURE: 0.967 seconds (recording time).\newline Describe the robot's visible progress on this task and its likely next movement.''}{The robot is beside a light-gray tabletop supported by a dark frame. Its right hand is extended toward a hammer with a yellow-brown handle and a dark metal-looking head, while the left hand remains away from the object.}{The robot may refine the right-hand alignment and close its fingers around the handle before lifting the hammer from the desk.}{}
\par\vspace{5pt}
\lfappbexamplecard{}{\includegraphics[width=\linewidth,height=1.35in,keepaspectratio]{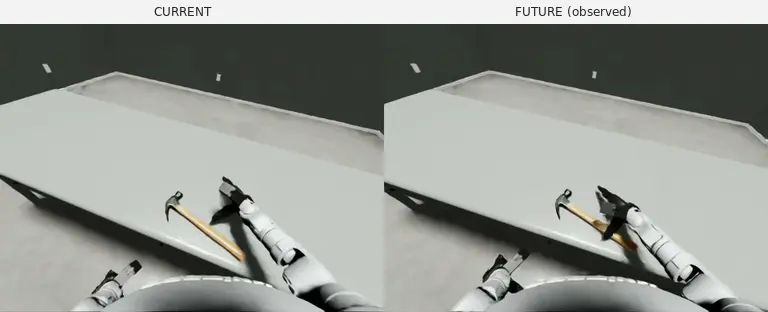}}{\textbf{System prompt}\newline ``Give a grounded account of humanoid behavior, including posture, object interaction, and visible appearance details. Use Task progress: for the observed task state, followed by Future prediction: for a plausible next action. Write two paragraphs, avoid image references, and qualify uncertain claims.''\par\vspace{2pt}\textbf{User prompt}\newline ``Task context: Move the hammer from the source desk to the basket on the other desk.\newline Elapsed time from CURRENT to FUTURE: 0.967 seconds (recording time).\newline Describe the robot's visible progress on this task and its likely next movement.''}{The robot's right hand is positioned around the hammer handle near the edge of the pale tabletop. The tool remains close to the surface, and the left arm stays clear; the visible state suggests grasp alignment or initial acquisition.}{The robot may secure the hammer and raise its right hand, keeping the tool clear of the desk edge before turning toward the receiving desk.}{}
\par\vspace{5pt}
\clearpage
\noindent\textbf{DoubleDesk}\par\vspace{3pt}
\lfappbexamplecard{}{\includegraphics[width=\linewidth,height=1.35in,keepaspectratio]{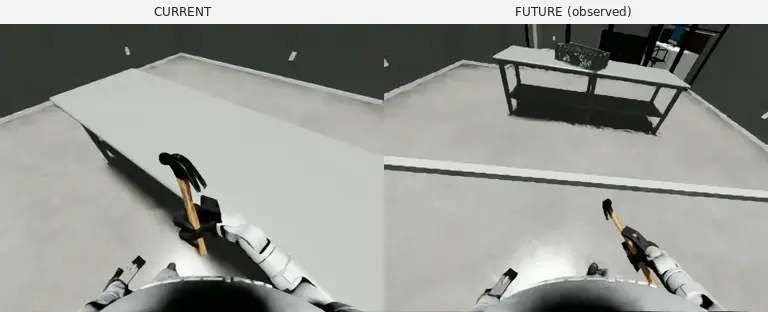}}{\textbf{System prompt}\newline ``Explain what the robot is visibly doing and how its behavior may continue. Include relevant object colors, textures, and apparent materials. Start the two paragraphs with Task progress: and Future prediction:, respectively. Focus on the robot and distinguish observed behavior from uncertain prediction.''\par\vspace{2pt}\textbf{User prompt}\newline ``Task context: Move the hammer from the source desk to the basket on the other desk.\newline Elapsed time from CURRENT to FUTURE: 0.967 seconds (recording time).\newline Describe the robot's visible progress on this task and its likely next movement.''}{The robot is oriented toward a second pale desk carrying a dark perforated basket. The hammer is visible near its right hand below the tabletop, while the source desk is no longer directly in front of the robot.}{The robot is likely to approach the receiving desk and raise the hammer toward the open basket, adjusting its body position for placement.}{}
\par\vspace{5pt}
\lfappbexamplecard{}{\includegraphics[width=\linewidth,height=1.35in,keepaspectratio]{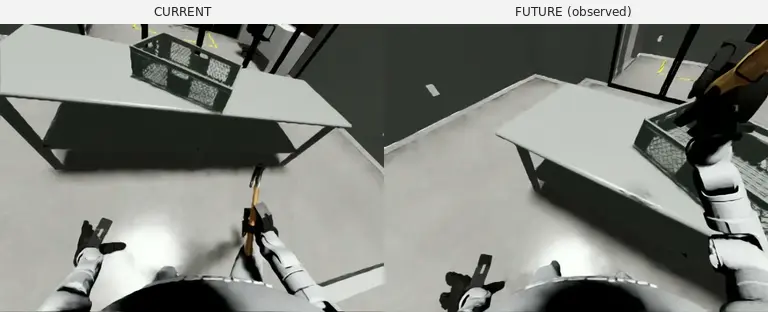}}{\textbf{System prompt}\newline ``Describe task progress through the humanoid's pose, movement, and relationship to nearby objects. Mention useful color and material cues. Then provide a cautious near-term continuation. Use exactly two paragraphs, headed Task progress: and Future prediction:, without narrating individual images or frames.''\par\vspace{2pt}\textbf{User prompt}\newline ``Task context: Move the hammer from the source desk to the basket on the other desk.\newline Elapsed time from CURRENT to FUTURE: 0.967 seconds (recording time).\newline Describe the robot's visible progress on this task and its likely next movement.''}{The robot is close to the dark mesh-like basket on the receiving desk. Its right hand and the yellow-brown-handled hammer are positioned at the basket opening, with the forearm extending toward the container; release is not visibly established.}{The robot may lower the hammer into the basket and release its grip, then withdraw its right hand once the tool is supported.}{}
\par\vspace{5pt}
\clearpage
\noindent\textbf{P\&PBox}\par\vspace{3pt}
\lfappbexamplecard{}{\includegraphics[width=\linewidth,height=1.35in,keepaspectratio]{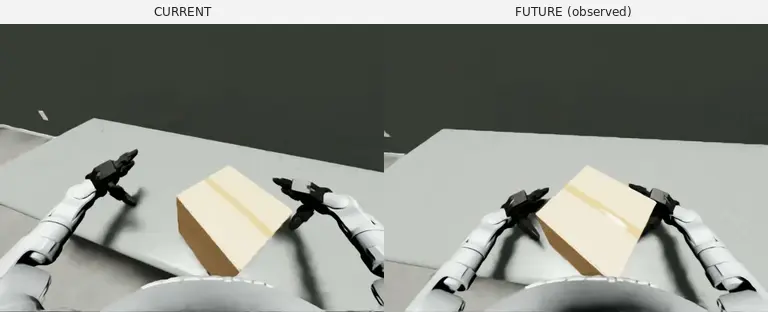}}{\textbf{System prompt}\newline ``Describe the humanoid's visible state, actions, spatial relations, and relevant colors and apparent materials. Write two paragraphs headed Task progress: and Future prediction:, in that order. Keep predictions tentative and grounded in observed behavior; describe the robot without referring to images or frames.''\par\vspace{2pt}\textbf{User prompt}\newline ``Task context: Pick up the box and place it on the shelf.\newline Elapsed time from CURRENT to FUTURE: 0.967 seconds (recording time).\newline Describe the robot's visible progress on this task and its likely next movement.''}{The robot is facing a tan cardboard-looking box with a pale taped seam on a light-gray desk. Both black hands are extended toward opposite sides of the box, with the forearms angled inward as the robot aligns for a two-handed grasp.}{The robot may bring both hands into contact with the box and secure it before attempting to lift it from the tabletop.}{}
\par\vspace{5pt}
\lfappbexamplecard{}{\includegraphics[width=\linewidth,height=1.35in,keepaspectratio]{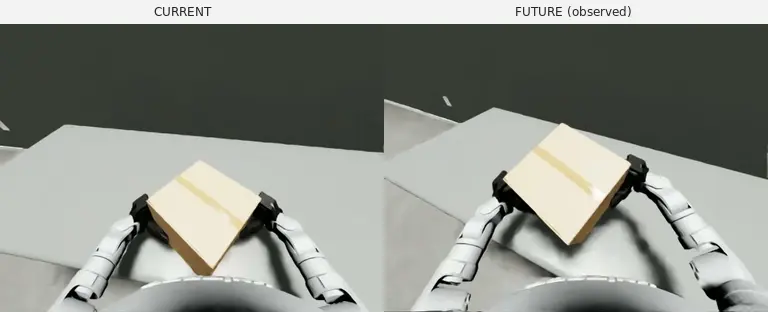}}{\textbf{System prompt}\newline ``Give a grounded account of humanoid behavior, including posture, object interaction, and visible appearance details. Use Task progress: for the observed task state, followed by Future prediction: for a plausible next action. Write two paragraphs, avoid image references, and qualify uncertain claims.''\par\vspace{2pt}\textbf{User prompt}\newline ``Task context: Pick up the box and place it on the shelf.\newline Elapsed time from CURRENT to FUTURE: 0.967 seconds (recording time).\newline Describe the robot's visible progress on this task and its likely next movement.''}{The tan box is positioned between the robot's black hands, with both arms held forward near the pale desk. The robot is turning away from the desk while maintaining the box between its hands; the exact support forces are not visible.}{The robot is likely to continue orienting toward the storage rack while keeping the box close and supported between its hands.}{}
\par\vspace{5pt}
\clearpage
\noindent\textbf{P\&PBox}\par\vspace{3pt}
\lfappbexamplecard{}{\includegraphics[width=\linewidth,height=1.35in,keepaspectratio]{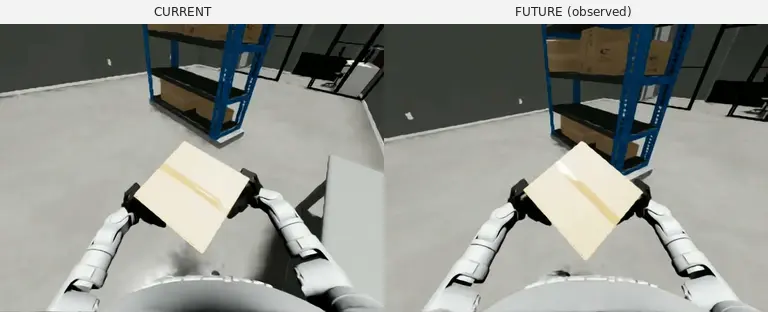}}{\textbf{System prompt}\newline ``Explain what the robot is visibly doing and how its behavior may continue. Include relevant object colors, textures, and apparent materials. Start the two paragraphs with Task progress: and Future prediction:, respectively. Focus on the robot and distinguish observed behavior from uncertain prediction.''\par\vspace{2pt}\textbf{User prompt}\newline ``Task context: Pick up the box and place it on the shelf.\newline Elapsed time from CURRENT to FUTURE: 0.967 seconds (recording time).\newline Describe the robot's visible progress on this task and its likely next movement.''}{The robot is carrying the tan box in front of its torso toward a blue-framed rack with dark shelves. Both hands remain at the sides of the box, and the open lower shelf is directly ahead.}{The robot may take a short step toward the rack and adjust the box height and orientation to align it with the available shelf space.}{}
\par\vspace{5pt}
\lfappbexamplecard{}{\includegraphics[width=\linewidth,height=1.35in,keepaspectratio]{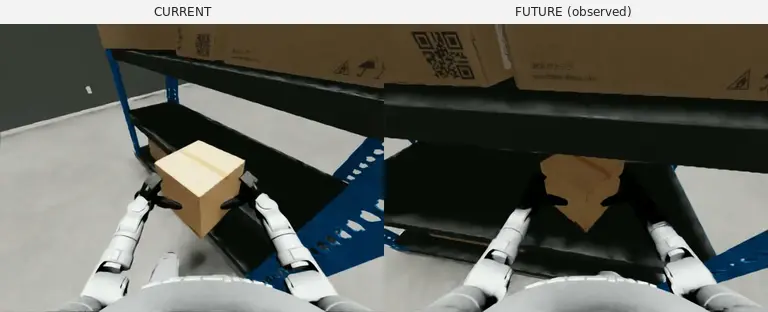}}{\textbf{System prompt}\newline ``Describe task progress through the humanoid's pose, movement, and relationship to nearby objects. Mention useful color and material cues. Then provide a cautious near-term continuation. Use exactly two paragraphs, headed Task progress: and Future prediction:, without narrating individual images or frames.''\par\vspace{2pt}\textbf{User prompt}\newline ``Task context: Pick up the box and place it on the shelf.\newline Elapsed time from CURRENT to FUTURE: 0.967 seconds (recording time).\newline Describe the robot's visible progress on this task and its likely next movement.''}{The robot's forearms extend toward the lower dark shelf of the blue rack, with the tan box positioned at the shelf opening. The box is partially obscured by the shelf structure, so a completed release cannot be confirmed.}{The robot may move the box farther onto the shelf, release it once supported, and withdraw both hands from the rack.}{}
\par\vspace{5pt}
\clearpage
\noindent\textbf{OpenDoor}\par\vspace{3pt}
\lfappbexamplecard{}{\includegraphics[width=\linewidth,height=1.35in,keepaspectratio]{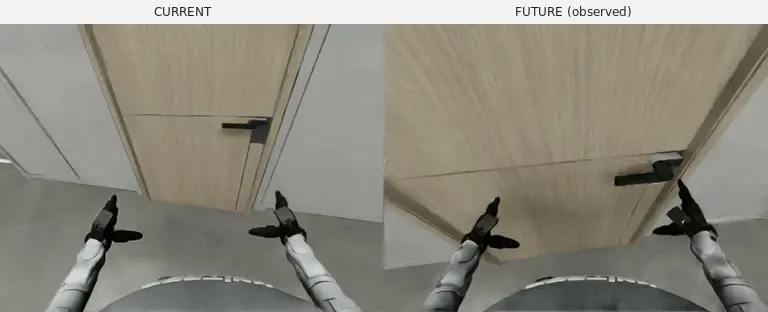}}{\textbf{System prompt}\newline ``Describe the humanoid's task-relevant state and movement in concrete visual terms, including relevant colors, textures, and apparent materials. Follow with a cautious next-action prediction. Write two paragraphs beginning Task progress: and Future prediction:, respectively; discuss robot behavior without naming images or frames.''\par\vspace{2pt}\textbf{User prompt}\newline ``Task context: Open the door.\newline Elapsed time from CURRENT to FUTURE: 0.967 seconds (recording time).\newline Describe how the robot is engaging with the door handle and how this behavior may continue.''}{The robot is close to a closed light-wood door with visible grain and a dark horizontal lever handle. Both arms are held forward, and the right hand is directed toward the handle while the left hand remains lower and farther away.}{The robot may raise and align its right hand with the lever, establishing contact before attempting to operate the handle.}{}
\par\vspace{5pt}
\lfappbexamplecard{}{\includegraphics[width=\linewidth,height=1.35in,keepaspectratio]{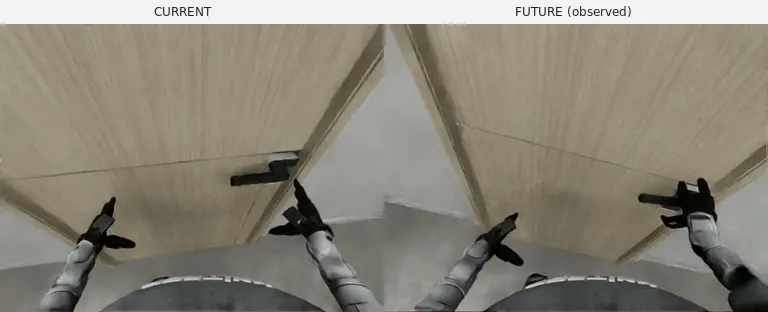}}{\textbf{System prompt}\newline ``Describe what is visibly happening around the robot and how it relates to the task, including relevant posture, object appearance, color, and apparent material. Follow with a tentative next step. Use Task progress: and Future prediction: in separate paragraphs without image-by-image narration.''\par\vspace{2pt}\textbf{User prompt}\newline ``Task context: Open the door.\newline Elapsed time from CURRENT to FUTURE: 0.967 seconds (recording time).\newline Characterize the visible task state around the robot and door, followed by a near-term forecast.''}{The robot's right black hand is positioned around the dark lever on the light-wood door, while its left hand stays clear of the handle. The door still occupies most of the space ahead, and the visible configuration suggests handle engagement rather than a completed opening.}{The robot may maintain contact and move the handle before drawing the door open; the release of any latch cannot be inferred from appearance alone.}{}
\par\vspace{5pt}
\clearpage
\noindent\textbf{OpenDoor}\par\vspace{3pt}
\lfappbexamplecard{}{\includegraphics[width=\linewidth,height=1.35in,keepaspectratio]{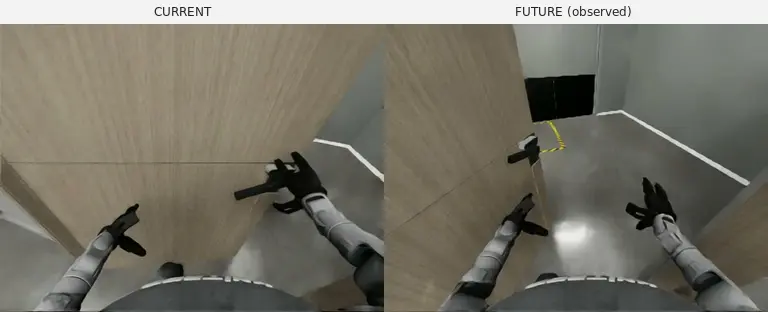}}{\textbf{System prompt}\newline ``Provide a concrete description of the humanoid's observed state and a cautious prediction of its next task-relevant action. Include spatial relationships and useful color, texture, or apparent-material details. Write Task progress: before Future prediction: in two paragraphs, without referring to images or frames.''\par\vspace{2pt}\textbf{User prompt}\newline ``Task context: Open the door.\newline Elapsed time from CURRENT to FUTURE: 0.967 seconds (recording time).\newline Describe current progress toward opening the door and a plausible next step grounded in the robot's state.''}{The light-wood door is partly open, revealing gray flooring and a yellow-bordered area beyond the doorway. The robot's right hand is no longer wrapped around the dark lever, and both hands are held in front of the body near the opening.}{The robot may reposition its arms and body to maintain clearance from the moving door and approach the available passage.}{}
\par\vspace{5pt}
\lfappbexamplecard{}{\includegraphics[width=\linewidth,height=1.35in,keepaspectratio]{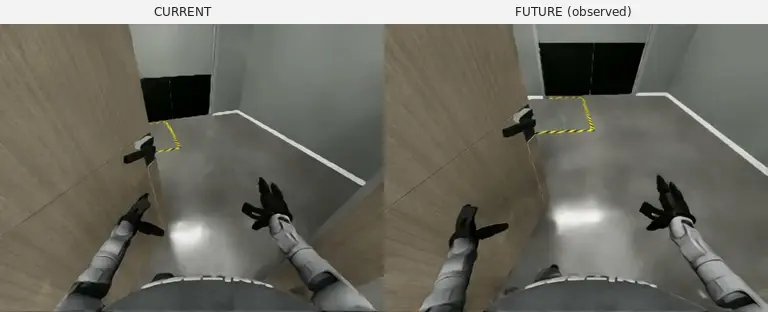}}{\textbf{System prompt}\newline ``Write a detailed account of the humanoid's task state, followed by a cautious forecast of its next action. Use the paragraph headings Task progress: and Future prediction:. Ground the description in visible body and object states, include appearance details, and avoid image references.''\par\vspace{2pt}\textbf{User prompt}\newline ``Task context: Open the door.\newline Elapsed time from CURRENT to FUTURE: 0.967 seconds (recording time).\newline Describe the visible robot, handle, and door configuration, then give a near-term prediction.''}{The robot is beside the partly open light-wood door, with the dark handle to its left and a wider view of the gray room beyond. Both forearms are held forward and clear of the handle, while the yellow-bordered floor area is visible ahead.}{The robot may move toward the doorway while checking clearance around the door edge; crossing into the next room remains a possible continuation rather than an observed outcome.}{}
\par\vspace{5pt}
\clearpage
\noindent\textbf{SitSofa}\par\vspace{3pt}
\lfappbexamplecard{}{\includegraphics[width=\linewidth,height=1.35in,keepaspectratio]{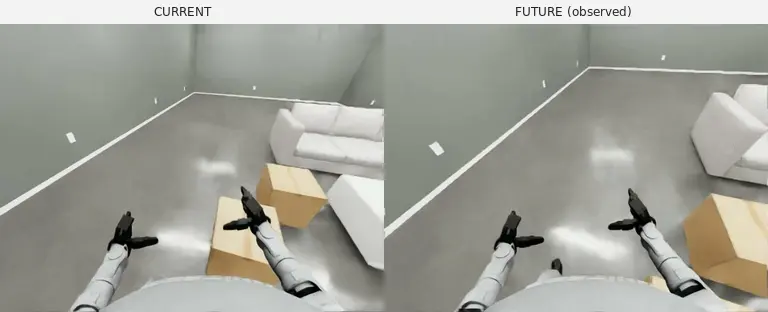}}{\textbf{System prompt}\newline ``Describe the humanoid's visible state, actions, spatial relations, and relevant colors and apparent materials. Write two paragraphs headed Task progress: and Future prediction:, in that order. Keep predictions tentative and grounded in observed behavior; describe the robot without referring to images or frames.''\par\vspace{2pt}\textbf{User prompt}\newline ``Task context: Approach the sofa and sit down.\newline Elapsed time from CURRENT to FUTURE: 0.967 seconds (recording time).\newline Describe the robot's visible progress on this task and its likely next movement.''}{The robot is moving toward an off-white upholstered sofa beside a low white rectangular table. Tan cardboard-looking boxes occupy part of the gray floor ahead, and both arms are held forward as the robot positions itself near the furniture.}{The robot may adjust its path around the boxes and table to approach the open side of the sofa.}{}
\par\vspace{5pt}
\lfappbexamplecard{}{\includegraphics[width=\linewidth,height=1.35in,keepaspectratio]{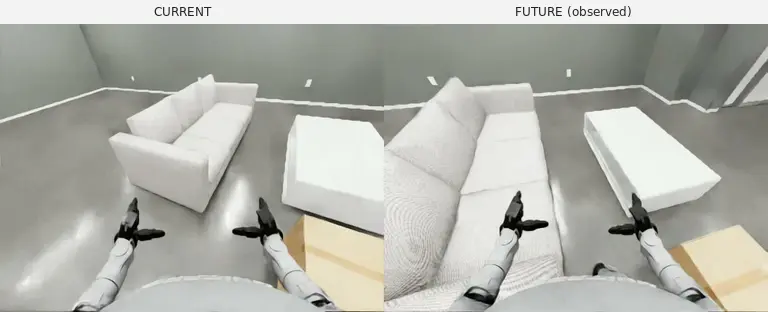}}{\textbf{System prompt}\newline ``Give a grounded account of humanoid behavior, including posture, object interaction, and visible appearance details. Use Task progress: for the observed task state, followed by Future prediction: for a plausible next action. Write two paragraphs, avoid image references, and qualify uncertain claims.''\par\vspace{2pt}\textbf{User prompt}\newline ``Task context: Approach the sofa and sit down.\newline Elapsed time from CURRENT to FUTURE: 0.967 seconds (recording time).\newline Describe the robot's visible progress on this task and its likely next movement.''}{The robot is next to the off-white sofa, with its textured cushions on one side and the white low table on the other. Its hands remain in front of the torso as it aligns within the narrow space beside the seat.}{The robot may turn and reposition its feet to align its body with the sofa before lowering toward the cushion.}{}
\par\vspace{5pt}
\clearpage
\noindent\textbf{SitSofa}\par\vspace{3pt}
\lfappbexamplecard{}{\includegraphics[width=\linewidth,height=1.35in,keepaspectratio]{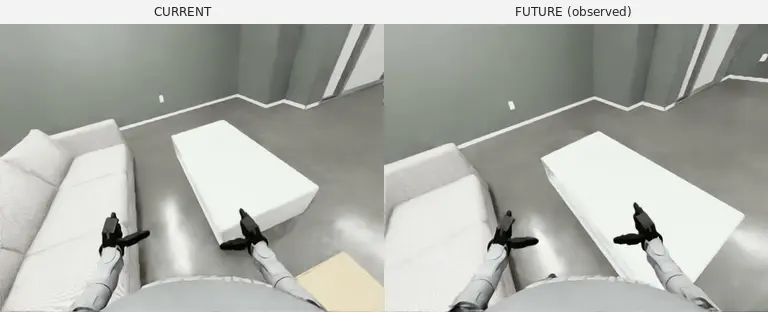}}{\textbf{System prompt}\newline ``Explain what the robot is visibly doing and how its behavior may continue. Include relevant object colors, textures, and apparent materials. Start the two paragraphs with Task progress: and Future prediction:, respectively. Focus on the robot and distinguish observed behavior from uncertain prediction.''\par\vspace{2pt}\textbf{User prompt}\newline ``Task context: Approach the sofa and sit down.\newline Elapsed time from CURRENT to FUTURE: 0.967 seconds (recording time).\newline Describe the robot's visible progress on this task and its likely next movement.''}{The robot is very close to the sofa's pale textured seat cushions, with the white table directly ahead. Its forearms remain forward while its orientation changes near the seat; hip-seat contact is outside the visible region.}{The robot may finish aligning its body and begin lowering onto the sofa while keeping its arms clear of the table.}{}
\par\vspace{5pt}
\lfappbexamplecard{}{\includegraphics[width=\linewidth,height=1.35in,keepaspectratio]{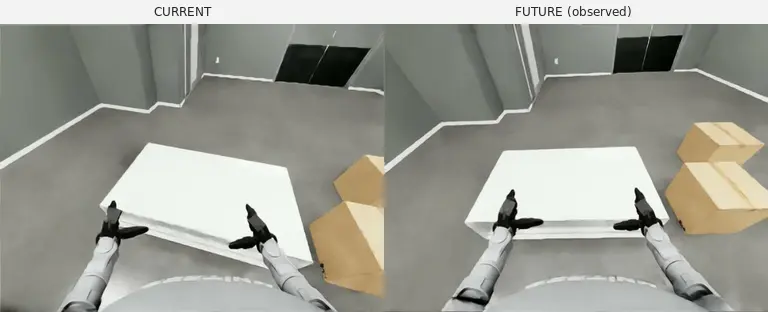}}{\textbf{System prompt}\newline ``Describe task progress through the humanoid's pose, movement, and relationship to nearby objects. Mention useful color and material cues. Then provide a cautious near-term continuation. Use exactly two paragraphs, headed Task progress: and Future prediction:, without narrating individual images or frames.''\par\vspace{2pt}\textbf{User prompt}\newline ``Task context: Approach the sofa and sit down.\newline Elapsed time from CURRENT to FUTURE: 0.967 seconds (recording time).\newline Describe the robot's visible progress on this task and its likely next movement.''}{The robot faces the low white table at close range, with the off-white sofa beside its body and tan boxes beyond the table. Its arms remain forward in a compact posture; the egocentric view alone does not establish whether the pelvis is supported by the cushion.}{The robot may settle into a stable seated posture and relax its arms if seat contact has been established, or make a small positioning adjustment if it has not.}{}
\par\vspace{5pt}
\clearpage
\noindent\textbf{Boxing}\par\vspace{3pt}
\lfappbexamplecard{}{\includegraphics[width=\linewidth,height=1.35in,keepaspectratio]{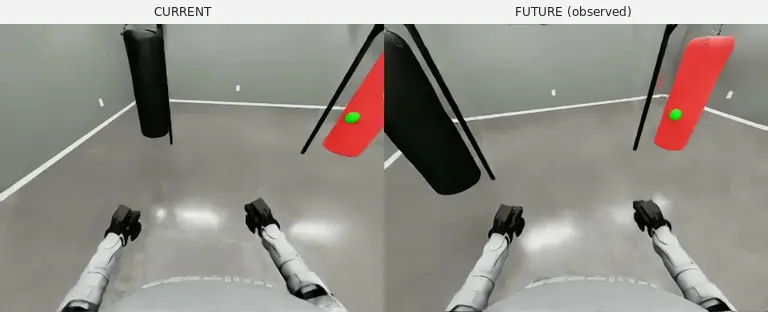}}{\textbf{System prompt}\newline ``Describe the humanoid's visible state, actions, spatial relations, and relevant colors and apparent materials. Write two paragraphs headed Task progress: and Future prediction:, in that order. Keep predictions tentative and grounded in observed behavior; describe the robot without referring to images or frames.''\par\vspace{2pt}\textbf{User prompt}\newline ``Task context: Reach and strike the marked target on the punching bag.\newline Elapsed time from CURRENT to FUTURE: 0.967 seconds (recording time).\newline Describe the robot's visible progress on this task and its likely next movement.''}{The robot is approaching a red cylindrical punching bag with a bright green target marker. A second dark bag hangs nearby, while both black hands are held in front of the robot at a distance from the marked target.}{The robot may move closer and orient its torso toward the green marker before extending a striking arm.}{}
\par\vspace{5pt}
\lfappbexamplecard{}{\includegraphics[width=\linewidth,height=1.35in,keepaspectratio]{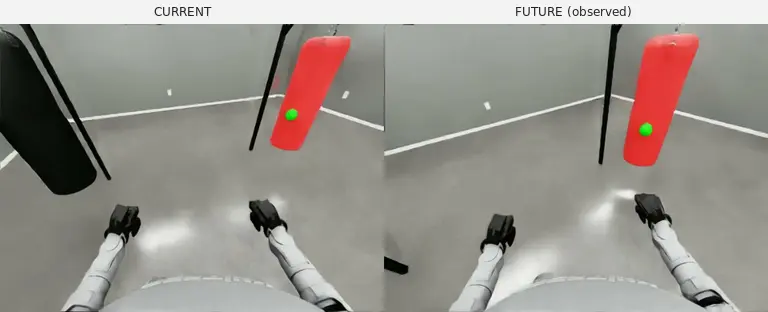}}{\textbf{System prompt}\newline ``Give a grounded account of humanoid behavior, including posture, object interaction, and visible appearance details. Use Task progress: for the observed task state, followed by Future prediction: for a plausible next action. Write two paragraphs, avoid image references, and qualify uncertain claims.''\par\vspace{2pt}\textbf{User prompt}\newline ``Task context: Reach and strike the marked target on the punching bag.\newline Elapsed time from CURRENT to FUTURE: 0.967 seconds (recording time).\newline Describe the robot's visible progress on this task and its likely next movement.''}{The red punching bag is more directly in front of the robot, with its green circular marker clearly visible. Both forearms remain raised ahead of the torso, and the robot is within a shorter approach distance of the target.}{The robot is likely to refine its stance and bring its right hand toward the marked area while keeping the other arm clear.}{}
\par\vspace{5pt}
\clearpage
\noindent\textbf{Boxing}\par\vspace{3pt}
\lfappbexamplecard{}{\includegraphics[width=\linewidth,height=1.35in,keepaspectratio]{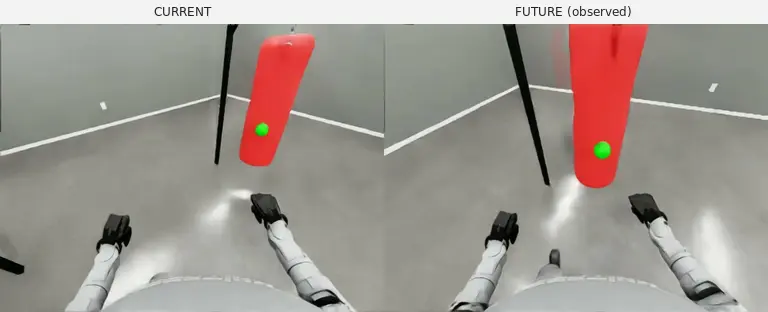}}{\textbf{System prompt}\newline ``Explain what the robot is visibly doing and how its behavior may continue. Include relevant object colors, textures, and apparent materials. Start the two paragraphs with Task progress: and Future prediction:, respectively. Focus on the robot and distinguish observed behavior from uncertain prediction.''\par\vspace{2pt}\textbf{User prompt}\newline ``Task context: Reach and strike the marked target on the punching bag.\newline Elapsed time from CURRENT to FUTURE: 0.967 seconds (recording time).\newline Describe the robot's visible progress on this task and its likely next movement.''}{The robot is close to the red bag, with its right forearm directed toward the green target and its left hand held lower to the side. The bag's smooth red surface occupies much of the space ahead, but contact is not yet clearly established.}{The robot may extend the right arm farther to strike the green marker and then begin retracting the hand.}{}
\par\vspace{5pt}
\lfappbexamplecard{}{\includegraphics[width=\linewidth,height=1.35in,keepaspectratio]{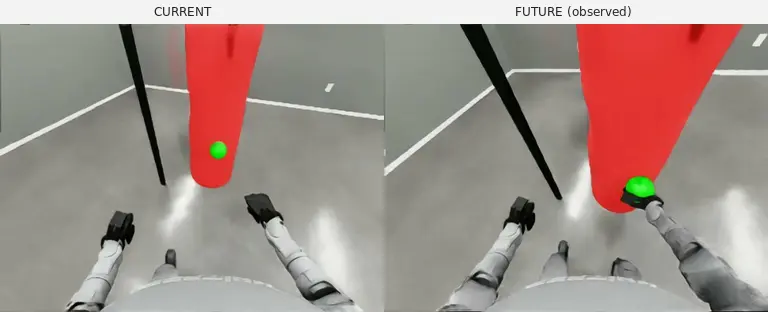}}{\textbf{System prompt}\newline ``Describe task progress through the humanoid's pose, movement, and relationship to nearby objects. Mention useful color and material cues. Then provide a cautious near-term continuation. Use exactly two paragraphs, headed Task progress: and Future prediction:, without narrating individual images or frames.''\par\vspace{2pt}\textbf{User prompt}\newline ``Task context: Reach and strike the marked target on the punching bag.\newline Elapsed time from CURRENT to FUTURE: 0.967 seconds (recording time).\newline Describe the robot's visible progress on this task and its likely next movement.''}{The robot's right black hand is extended against or immediately adjacent to the green marker on the red punching bag. The left hand remains away from the target, and the right arm configuration is consistent with the contact phase of a strike.}{The robot may retract its right arm and return to a balanced ready posture while the suspended bag may move in response to the interaction.}{}
\par\vspace{5pt}
\clearpage
\noindent\textbf{VisNavi}\par\vspace{3pt}
\lfappbexamplecard{}{\includegraphics[width=\linewidth,height=1.35in,keepaspectratio]{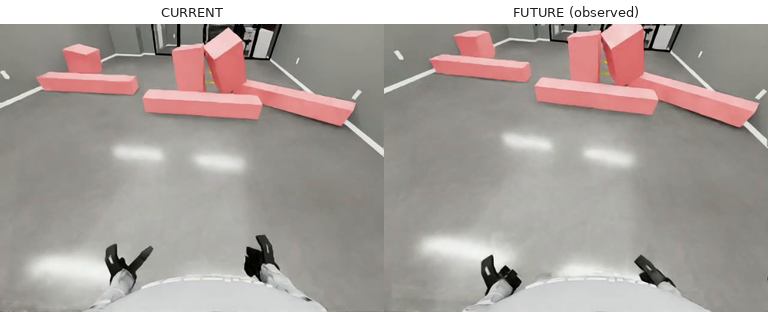}}{\textbf{System prompt}\newline ``Describe the humanoid's visible state, actions, spatial relations, and relevant colors and apparent materials. Write two paragraphs headed Task progress: and Future prediction:, in that order. Keep predictions tentative and grounded in observed behavior; describe the robot without referring to images or frames.''\par\vspace{2pt}\textbf{User prompt}\newline ``Task context: Navigate to the marked target while avoiding obstacles.\newline Elapsed time from CURRENT to FUTURE: 0.967 seconds (recording time).\newline Describe the robot's approach to the obstacles and its likely next movement toward the goal.''}{The robot is advancing across a smooth gray floor toward a cluster of matte pink rectangular obstacles. Its light-gray forearms and black hands are held forward, with a low horizontal beam ahead and taller blocks beyond it. The open floor narrows near the obstacles as the robot approaches the route through them.}{The robot may adjust its heading toward the gap between the pink barriers and continue forward while keeping its body clear of their edges.}{}
\par\vspace{5pt}
\lfappbexamplecard{}{\includegraphics[width=\linewidth,height=1.35in,keepaspectratio]{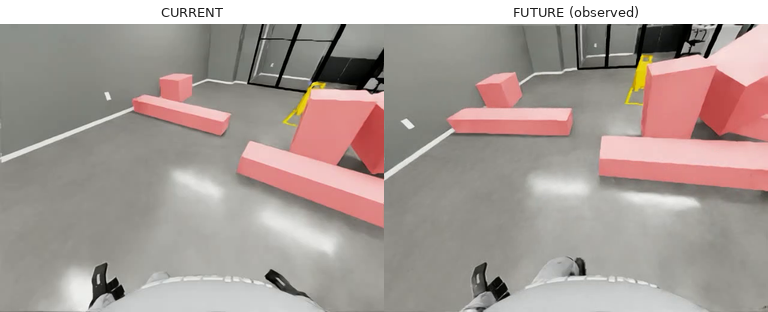}}{\textbf{System prompt}\newline ``Give a grounded account of humanoid behavior, including posture, object interaction, and visible appearance details. Use Task progress: for the observed task state, followed by Future prediction: for a plausible next action. Write two paragraphs, avoid image references, and qualify uncertain claims.''\par\vspace{2pt}\textbf{User prompt}\newline ``Task context: Navigate to the marked target while avoiding obstacles.\newline Elapsed time from CURRENT to FUTURE: 0.967 seconds (recording time).\newline Explain how the robot is moving through the obstacle field and where it may go next.''}{The robot is moving into the space between a low pink beam on its left and taller pink blocks on its right. Its black left hand remains low in front of the body as it adjusts its heading through the gap. Part of the yellow target marking is visible beyond the barriers, near the black-framed glass doors.}{The robot may continue through the gap, then turn toward the yellow marker while maintaining clearance from the low beam and taller blocks.}{}
\par\vspace{5pt}
\clearpage
\noindent\textbf{VisNavi}\par\vspace{3pt}
\lfappbexamplecard{}{\includegraphics[width=\linewidth,height=1.35in,keepaspectratio]{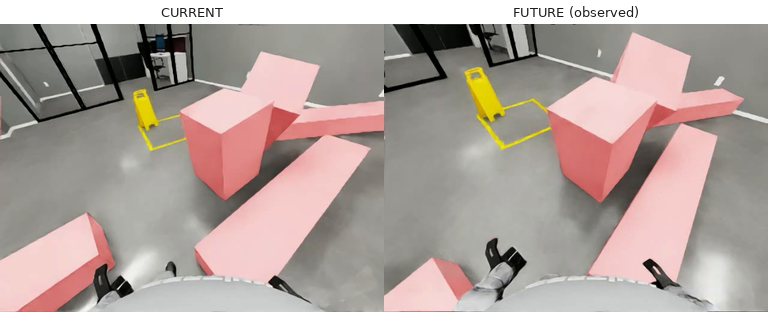}}{\textbf{System prompt}\newline ``Explain what the robot is visibly doing and how its behavior may continue. Include relevant object colors, textures, and apparent materials. Start the two paragraphs with Task progress: and Future prediction:, respectively. Focus on the robot and distinguish observed behavior from uncertain prediction.''\par\vspace{2pt}\textbf{User prompt}\newline ``Task context: Navigate to the marked target while avoiding obstacles.\newline Elapsed time from CURRENT to FUTURE: 0.967 seconds (recording time).\newline Describe the robot's position relative to nearby barriers and its probable next navigation action.''}{The robot is navigating close to a large pink cuboid on its right, beside a low horizontal beam. Its light-gray forearms remain forward and its black hands stay low as it turns around the obstruction. A yellow-outlined goal area and an upright yellow folding sign are visible ahead to the left of the block.}{The robot may pass along the open left side of the cuboid and approach the yellow boundary without brushing the nearby beam.}{}
\par\vspace{5pt}
\lfappbexamplecard{}{\includegraphics[width=\linewidth,height=1.35in,keepaspectratio]{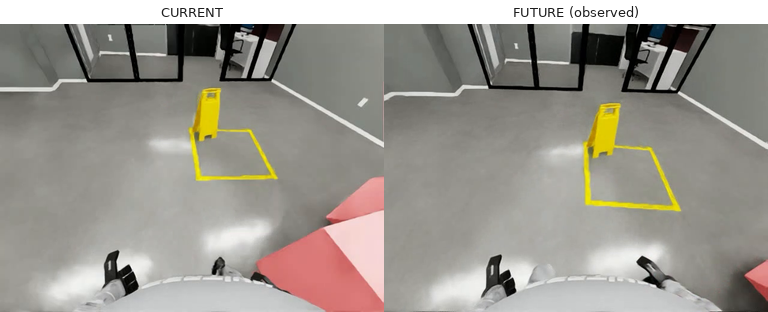}}{\textbf{System prompt}\newline ``Describe task progress through the humanoid's pose, movement, and relationship to nearby objects. Mention useful color and material cues. Then provide a cautious near-term continuation. Use exactly two paragraphs, headed Task progress: and Future prediction:, without narrating individual images or frames.''\par\vspace{2pt}\textbf{User prompt}\newline ``Task context: Navigate to the marked target while avoiding obstacles.\newline Elapsed time from CURRENT to FUTURE: 0.967 seconds (recording time).\newline Summarize the robot's progress toward the marked target and predict its immediate continuation.''}{The robot is emerging from the pink obstacle cluster onto clear gray flooring. The yellow rectangular target boundary is directly ahead, with a yellow folding sign near its far edge and black-framed glass doors beyond. Both black hands remain forward while the nearest pink block recedes to the right; the robot is still approaching the marked area.}{The robot may advance into the yellow-outlined area and reduce its forward motion as it reaches the navigation target.}{}
\par\vspace{5pt}
\endgroup

\clearpage
\begingroup
\raggedbottom

\section{Qualitative Rollouts and Failure Cases}
\label{sec:qualitative_rollouts}

Figures~\ref{fig:appc_sit_sofa}--\ref{fig:appc_visual_navigation}
present selected egocentric rollouts for the seven HumanoidArena tasks
described in Section~\ref{sec:data_protocol}.
We show two successful rollouts and one failed rollout per task,
giving twenty-one sequences in total.
Each montage contains ten frames in temporal order, read from left to
right across the top row and then across the bottom row.
Frames are selected at nonuniform intervals to capture task-relevant
transitions, and the original image colors are preserved.
Outcome labels follow the source recordings; the displayed frames need
not include the exact instant at which a success criterion is satisfied.
The failure descriptions identify observable execution failures, and
these selected examples do not estimate their frequency.

\newsavebox{\appcCaptionBox}
\newlength{\appcAvailableHeight}
\newlength{\appcImageHeight}

% Image and editable panel label.
\newcommand{\appcRollout}[2]{%
    \includegraphics[
        width=\linewidth,
        height=\appcImageHeight,
        keepaspectratio
    ]{figures/appendix/#1}\par
    \vspace{2pt}%
    {\small #2\par}%
}

\makeatletter
% Arguments:
% #1 Optional task introduction
% #2 Task heading
% #3 Image filename prefix
% #4 Figure label
% #5 Figure caption
\newcommand{\appcTaskPage}[5][]{%
    \par
    \begingroup
    \setlength{\parindent}{0pt}%
    \setlength{\parskip}{0pt}%

    % Task heading and optional reading guide.
    \noindent{\bfseries #2\par}%
    \if\relax\detokenize{#1}\relax
    \else
        \vspace{4pt}%
        \noindent #1\par
    \fi
    \vspace{8pt}%

    % Typeset the caption once and measure its height.
    \sbox{\appcCaptionBox}{%
        \begin{minipage}{\linewidth}
            \def\@captype{figure}%
            \setlength{\abovecaptionskip}{6pt}%
            \setlength{\belowcaptionskip}{0pt}%
            \caption{#5}%
            \label{#4}%
        \end{minipage}%
    }%

    % Determine the space remaining after the heading and introduction.
    \setlength{\appcAvailableHeight}{\textheight}%
    \ifdim\pagegoal<\maxdimen
        \setlength{\appcAvailableHeight}{%
            \dimexpr\pagegoal-\pagetotal\relax
        }%
    \fi
    \addtolength{\appcAvailableHeight}{-\baselineskip}%

    % Limit image height while preserving normal text sizes.
    \setlength{\appcImageHeight}{%
        \dimexpr
            \appcAvailableHeight
            -\ht\appcCaptionBox
            -\dp\appcCaptionBox
            -4\baselineskip
            -24pt
        \relax
    }%
    \divide\appcImageHeight by 3\relax

    % Natural-height block: fixed gaps, with unused space below.
    \noindent
    \begin{minipage}[t]{\linewidth}
        \centering

        \appcRollout{#3_success_1.png}{(a) Success 1}

        \vspace{8pt}

        \appcRollout{#3_success_2.png}{(b) Success 2}

        \vspace{8pt}

        \appcRollout{#3_failure.png}{(c) Failure}

        \vspace{4pt}
        \noindent\usebox{\appcCaptionBox}\par
    \end{minipage}\par
    \endgroup
}
\makeatother

% ----------------------------------------------------------------------
% Sit Sofa
% ----------------------------------------------------------------------

\medskip

\appcTaskPage
    {Sit Sofa}
    {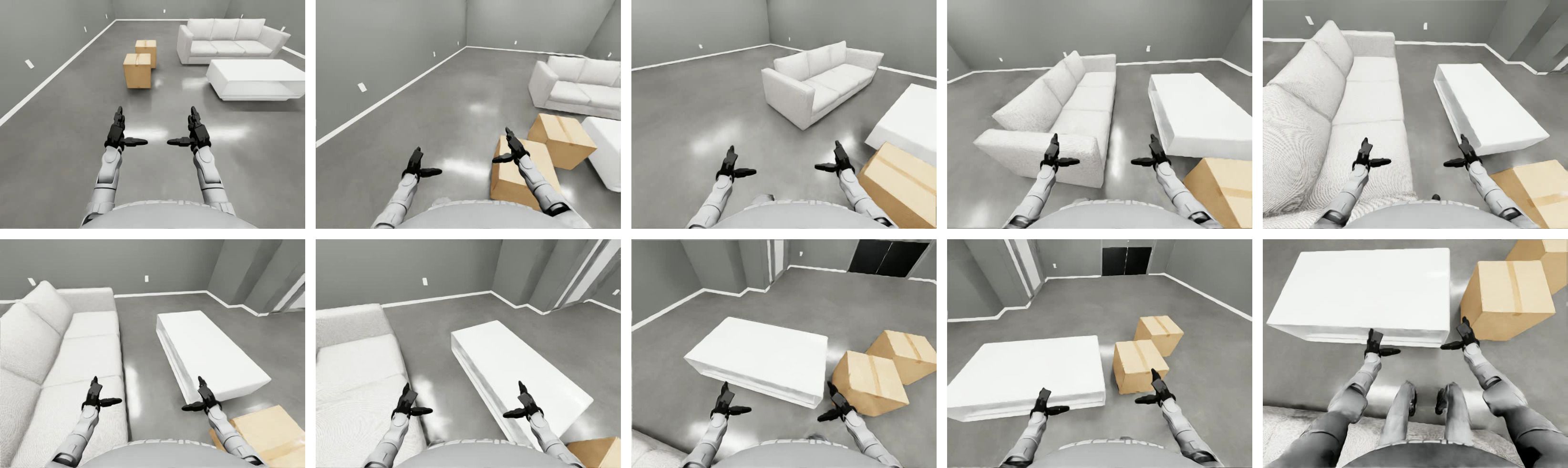}
    {fig:appc_sit_sofa}
    {Qualitative rollouts for Sit Sofa (SitSofa).
    Panels (a) and (b) show successful rollouts, including the approach
    to the sofa and reorientation for sitting.
    Panel (c) shows incomplete body lowering: the robot approaches the
    sofa but does not lower its body sufficiently to complete the
    sitting motion.}

\clearpage

% ----------------------------------------------------------------------
% Pick & Place Box
% ----------------------------------------------------------------------

\appcTaskPage[
    The robot must pick up the box and move it to the shelving unit.
    Samples (a) and (b) show successful pickup and transport,
    while (c) shows a failed pickup that leaves the box on the table.
]
    {Pick \& Place Box}
    {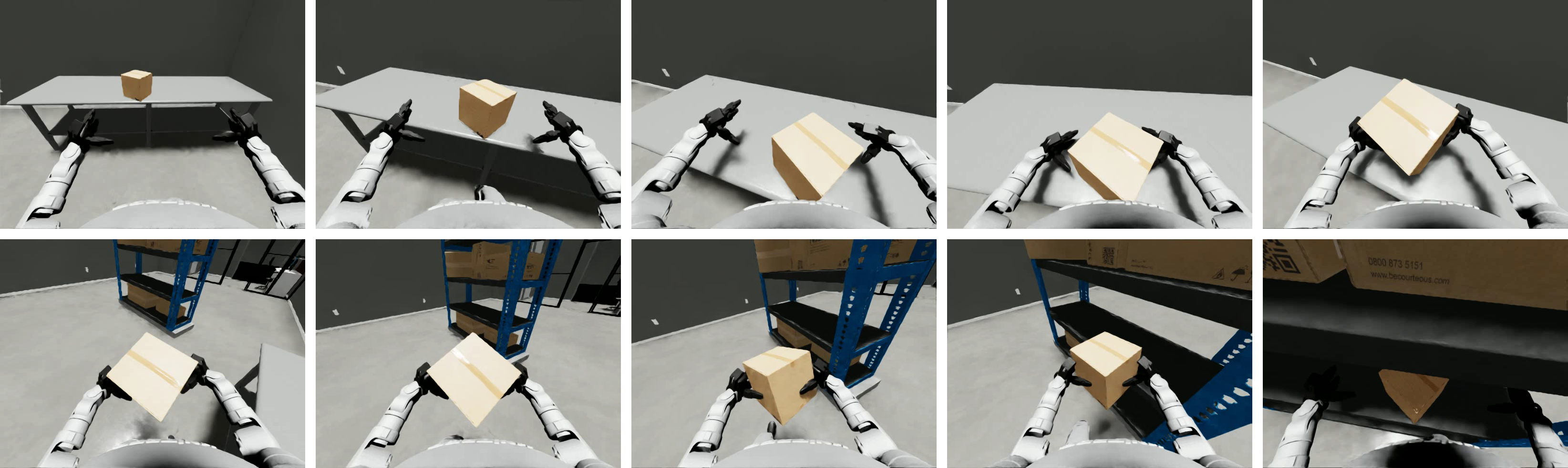}
    {fig:appc_pick_place_box}
    {Qualitative rollouts for Pick \& Place Box (P\&PBox).
    Panels (a) and (b) show successful rollouts, illustrating box pickup
    and transport toward the shelving unit.
    Panel (c) shows a failed pickup: the robot does not establish a
    grasp sufficient to lift the box, which remains on the table.}

\clearpage

% ----------------------------------------------------------------------
% Open Door
% ----------------------------------------------------------------------

\appcTaskPage[
    The robot must approach the door, operate the handle, and open it.
    Samples (a) and (b) show successful rollouts;
    in (c), the robot reaches toward the handle but the door remains closed.
]
    {Open Door}
    {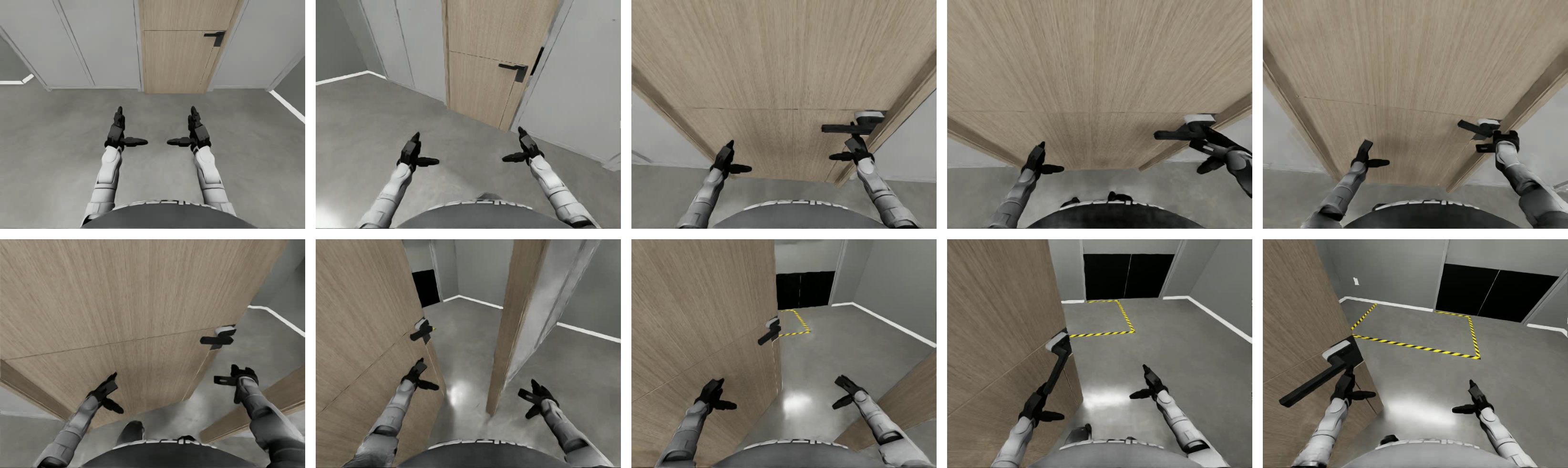}
    {fig:appc_open_door}
    {Qualitative rollouts for Open Door (OpenDoor).
    Panels (a) and (b) show successful rollouts, including approach,
    handle interaction, and door opening.
    Panel (c) shows unsuccessful handle operation: the robot reaches
    toward the handle but does not operate it sufficiently to open the
    door, which remains closed.}

\clearpage

% ----------------------------------------------------------------------
% Kick Football
% ----------------------------------------------------------------------

\appcTaskPage[
    The robot must approach the ball and kick it into the goal.
    Samples (a) and (b) show successful rollouts,
    while (c) shows a kick that misses the goal near the left post.
]
    {Kick Football}
    {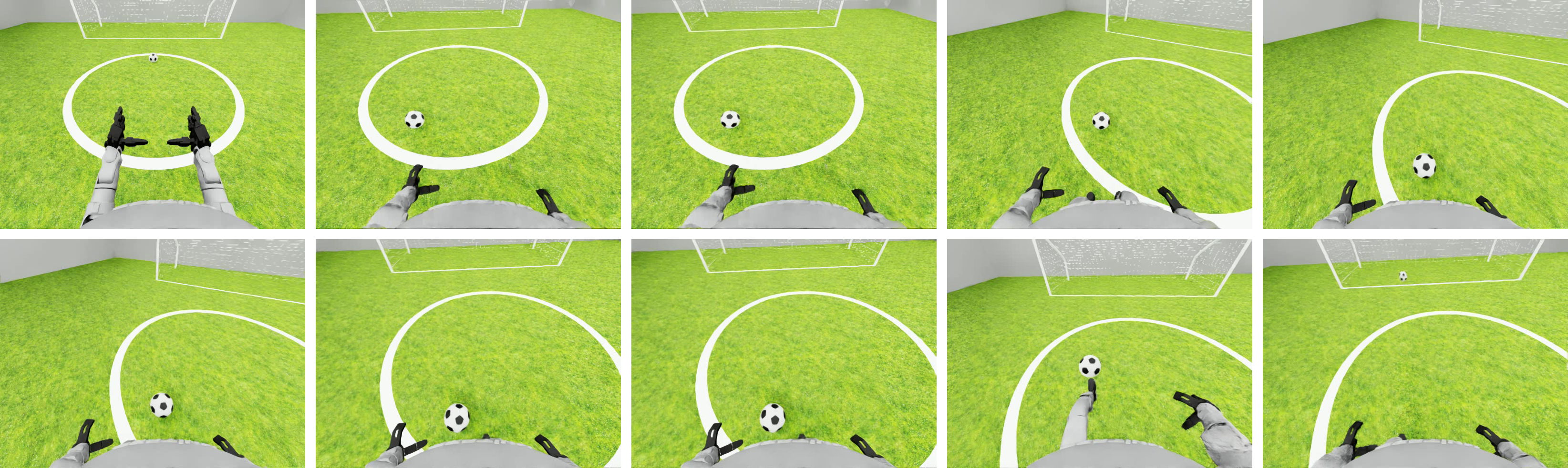}
    {fig:appc_kick_football}
    {Qualitative rollouts for Kick Football (Football).
    Panels (a) and (b) show successful rollouts, including the approach
    to the ball and the kick toward the goal.
    Panel (c) shows a missed goal: the ball travels toward the left
    goalpost and remains outside the goal.}

\clearpage

% ----------------------------------------------------------------------
% Double Desks Transfer
% ----------------------------------------------------------------------

\appcTaskPage[
    The robot must carry an object from one desk to the basket
    on the other desk.
    Samples (a) and (b) show successful transfers;
    in (c), the robot falls while carrying the object.
]
    {Double Desks Transfer}
    {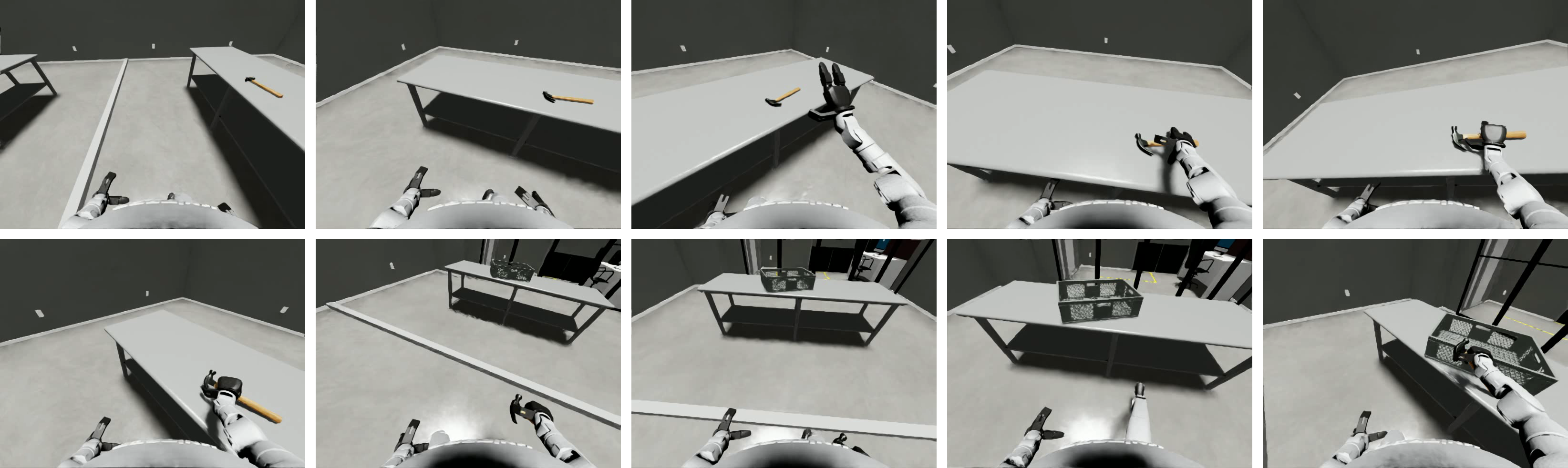}
    {fig:appc_double_desks}
    {Qualitative rollouts for Double Desks Transfer (DoubleDesk).
    Panels (a) and (b) show two successful rollouts involving pickup,
    transport, and transfer of the object between the desks.
    Panel (c) shows a fall during object transport: while carrying the
    object toward the receiving desk, the robot loses balance and
    falls before placing it in the basket.}

\clearpage

% ----------------------------------------------------------------------
% Boxing
% ----------------------------------------------------------------------

\appcTaskPage[
    The robot must strike the marked target on a punching bag.
    Samples (a) and (b) show successful rollouts,
    while (c) shows movement around the bags without a successful
    strike on the target.
]
    {Boxing}
    {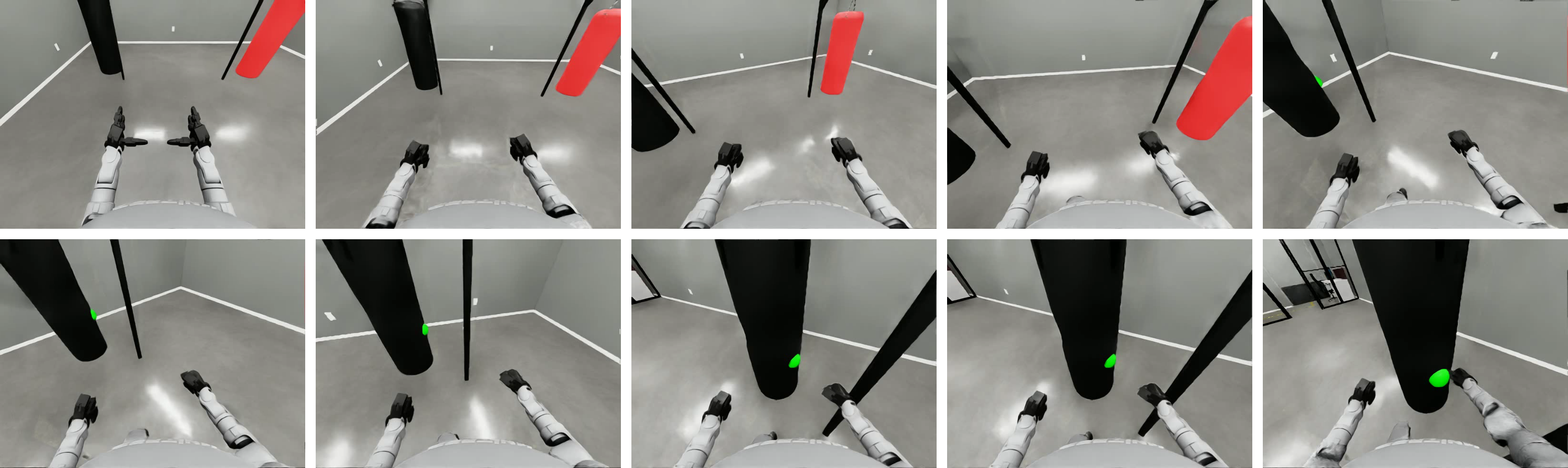}
    {fig:appc_boxing}
    {Qualitative rollouts for Boxing.
    Panels (a) and (b) show successful rollouts involving approach
    to the punching bags and target-directed arm motion.
    Panel (c) shows an unsuccessful target strike: the robot moves
    around the bags but does not land a punch on the marked target.}

\clearpage

% ----------------------------------------------------------------------
% Visual Navigation
% ----------------------------------------------------------------------

\appcTaskPage[
    The robot must navigate through obstacles to the marked goal region.
    Samples (a) and (b) show successful navigation;
    in (c), the robot contacts the obstacles and falls before
    reaching the goal.
]
    {Visual Navigation}
    {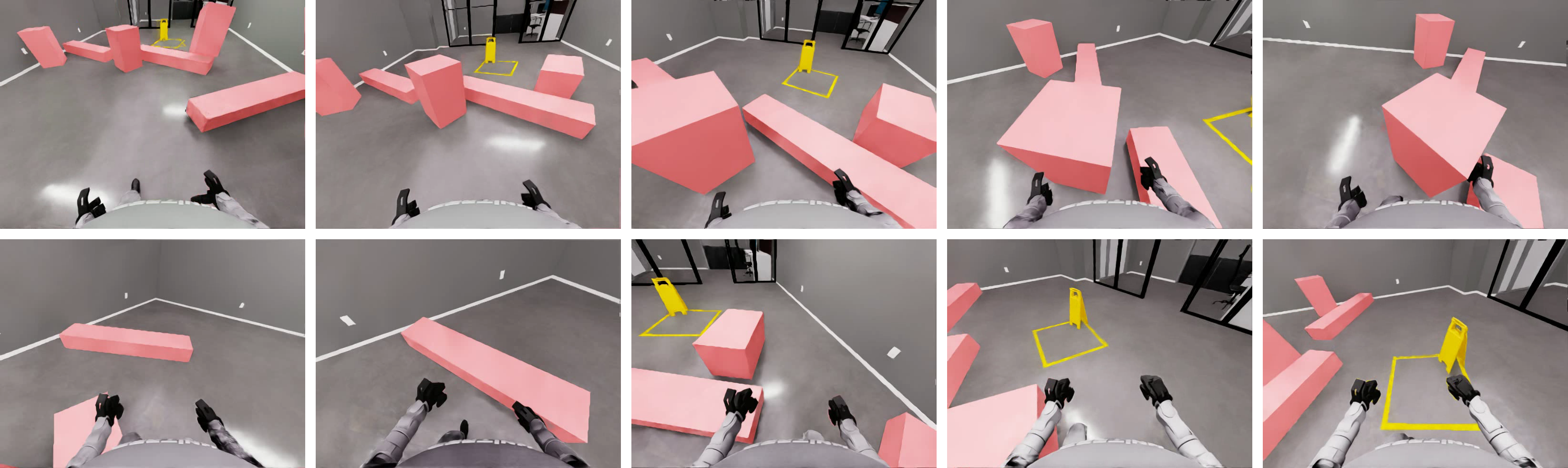}
    {fig:appc_visual_navigation}
    {Qualitative rollouts for Visual Navigation (VisNavi).
    Panels (a) and (b) show successful progress through the obstacle
    field toward the marked goal region.
    Panel (c) shows loss of balance during navigation: the robot
    contacts the obstacles and falls before reaching the goal.}

\clearpage
\endgroup

\end{document}